# Using Evidence-Centered Design to Guide Prompt Engineering in Measurement: An Illustration with Automated Essay Scoring

Jiawei Xiong
Curriculum Associates
Cheng Tang
The University of Georgia
Hong Jiao, Hanna Choi
The University of Maryland, College Park

## Abstract

Generative artificial intelligence (GenAI) models, particularly large language models (LLMs), are increasingly used in educational settings for tasks such as item generation, automated scoring, and feedback delivery. The quality of these outputs depends not only on the AI model but also on how users construct the prompts that guide model behaviors. Current prompt engineering practices are largely ad hoc, relying on trial and error rather than principled design. This lack of structure introduces validity threats including construct misrepresentation, construct-irrelevant variance, and limited reproducibility. We propose that Evidence-Centered Design (ECD), a framework in educational measurement for structuring valid inferences from observed performance, can serve as a principled foundation for prompt engineering in educational applications. We map the three core ECD models, the student model, the evidence model, and the task model, onto the components of educational prompt design, and describe a workflow for ECD-based prompt engineering. We derive five design principles from this mapping and present an empirical study contrasting ECD-designed prompts with state-of-the-art (SOTA) prompting baselines on an automated scoring task. Our empirical study demonstrates the usefulness of this framework on automated scoring. Results suggest that treating prompting as assessment design rather than informal instructions strengthens the validity, reliability, and fairness of AI-generated educational outputs.

## Introduction

Generative artificial intelligence (GenAI) has rapidly influenced assessment and education methods and practices. Large language models (LLMs) such as GPT and Claude can draft assessment items, score student responses, generate individualized feedback, and support lesson planning (Kasneci et al., 2023). Educators, students, and researchers now interact with these models routinely, integrating them into teaching, learning, and assessment workflows. This rapid adoption reflects genuine utility, because LLMs can perform at scale tasks that previously required substantial human labor. The quality of what these models produce depends on two factors. The obvious factor is the capability of the AI model itself, including its training data, architecture, and alignment procedures (Zubiaga, 2024). The less obvious but equally consequential factor is how users specify the task to the model (Federiakin et al., 2024). The instructions that users provide, known as prompts, shape model outputs in desired ways. Therefore, a vague prompt may yield a superficial or inaccurate response, while a carefully structured prompt may yield output that closely matches expert judgment. Because the same model can produce highly variable outputs depending on the prompt, the design of prompts is a central determinant of whether AI-generated educational outputs are accurate, aligned with learning objectives, and fair across learner populations.

Different prompting strategies have been developed to improve GenAI models' output quality, and each of these strategies has shown promise in improving model performance on various tasks, and practitioners often combine them. For example, zero-shot prompting provides the model with a task instruction and no examples, while few-shot prompting includes one or more demonstration examples in the prompt in addition to task instructions (Brown et al., 2020). Chain-of-thought prompting instructs the model to show intermediate reasoning steps before reaching a conclusion (Wei et al., 2022). Self-consistency prompting samples multiple reasoning paths and selects the most frequent answer (Wang et al., 2023). Role-based prompting assigns the model a persona or expert identity to shape its responses (Kong et al., 2024). Retrieval-augmented prompting supplements the prompt with retrieved documents to ground the model's output in specific source material (Lewis et al., 2020).

Despite the common use of these techniques, most prompt engineering in assessment and education settings remains naturalistic and ad hoc. That is, users typically write prompts based on intuition, iterating by trial and error until the output looks acceptable (Lee & Palmer, 2025). There is no systematic design process that connects what the user expects the model to produce with the structure of the prompt itself. This absence of principled design introduces several concrete problems. For instance, AI outputs may be inaccurate, containing factual errors or hallucinated content that the user does not detect (Sun et al., 2024). Additionally, the reasoning the model displays may be unclear or internally inconsistent, making it difficult to evaluate whether the output is trustworthy (Xie et al., 2024). In some cases, the model's stated reasoning may not align with the answer it provides, a form of deceptive behavior that undermines user trust (Turpin et al., 2024). Other than that, outputs may be weakly aligned with the learning objectives the user intended, addressing adjacent but different constructs (Küchemann et al., 2023). Another most troubling issue for educational applications is that outputs through ad hoc prompting could be difficult to reproduce across different users, contexts, or model versions, which limits their generalizability (Sclar et al., 2024).

These practical problems map directly onto validity concerns in educational measurement. On one hand, construct underrepresentation occurs when the prompt fails to elicit the full range of knowledge, skills, or abilities (KSAs) relevant to the target construct, producing outputs that are incomplete or shallow. On the other hand, construct-irrelevant variance occurs when the prompt introduces extraneous factors that influence the output but are unrelated to the construct of interest, such as response format preferences or surface-level stylistic cues. Also, unexamined biases may emerge when prompts are not designed with attention to how model behavior varies across learner subgroups, cultural contexts, or linguistic backgrounds. Beyond validity threats, pedagogical risks include misalignment with curricular standards, inaccurate feedback that reinforces misconceptions, improper identification of student errors, and inequitable performance for different learner populations. When prompts are not designed with explicit attention to these validity concerns, users have no principled basis to evaluate the trustworthiness of the outputs.

These challenges point to a fundamental gap in current practice. Before writing a prompt, users must articulate what they expect the model to produce, what evidence would indicate that the output is adequate, and how the prompt structure connects to those expectations. Without this articulation, prompt engineering remains an informal craft rather than a defensible design process. What is needed is a framework that connects intended inferences, evidence requirements, and task specifications in a coherent structure. Such a framework would allow users to design prompts that are not merely effective by accident but valid and reproducible by design.

We propose Evidence-Centered Design (ECD) as such a guiding framework. ECD has a long history in educational measurement as a principled approach to structuring valid inferences from observable behavior (Mislevy et al., 2003; Mislevy & Haertel, 2006). It was developed to ensure that every component of an assessment, from the task to the scoring rule to the reported score, can be traced back to an explicit argument about what is being measured and why the chosen evidence supports that measurement. We propose that the same logic applies directly to prompt engineering with principled design. A prompt is thus treated as an instrument or a performance assessment task designed to elicit a particular kind of output from an AI model or LLM performance, just as an assessment task is designed to elicit a particular kind of performance from a learner. If the prompt is not designed with attention to what it should produce, what counts as adequate evidence, and how the prompt structure supports those goals, then the resulting output cannot be defended as valid for its intended use.

## Background

### Prompt Engineering

At its simplest format, a prompt is a natural language instruction such as "Summarize this passage". In practice, prompts for assessment and education tasks are often more complex, incorporating contextual information, constraints on the output format, examples of desired responses, and evaluation criteria. Prompt engineering refers to the systematic design of these inputs to improve the quality of model outputs (Liu et al., 2023). A growing body of research has identified specific prompting techniques that reliably affect model behavior.

The simplest approach is zero-shot prompting, in which the user provides only the task instruction without any examples. Zero-shot prompting is convenient and widely used, but it provides the model with minimal guidance about the expected form, depth, or standards of the response (Kojima et al., 2022). It is more like intuitive, native responses from an AI model responding to the instruction. Few-shot prompting addresses this limitation by including one or more worked examples in the prompt, demonstrating the desired input-output mapping. Research has shown that few-shot examples can substantially improve model accuracy on structured tasks such as classification, scoring, and information extraction, because the examples anchor the model's behavior to a concrete standard (Brown et al., 2020).

Chain-of-thought (CoT) prompting takes a different approach by instructing the model to generate intermediate reasoning steps before producing a final answer, which has proven particularly effective for tasks that require multi-step reasoning, such as mathematical problem solving and logical inference (Wei et al., 2022). The underlying mechanism is that generating explicit reasoning steps constrains the model to produce answers that follow from stated premises. Self-consistency extends CoT prompting by sampling multiple reasoning paths and selecting the answer that appears most frequently across paths (Wang et al., 2023). This approach improves robustness by reducing the influence of any single erroneous reasoning chain.

Role-based prompting assigns the model a specific identity or expertise, such as "You are an experienced science teacher" or "You are an assessment specialist". Specifying a role activates patterns in the model's training data associated with that role, producing outputs that are more contextually appropriate (Kong et al., 2024). Retrieval-augmented prompting, sometimes called retrieval-augmented generation (RAG), supplements the prompt with documents retrieved from an external knowledge base, grounding the model's output in specific source material rather than relying solely on its parametric knowledge (Lewis et al., 2020). This technique is especially useful when accuracy depends on domain-specific or up-to-date information that the model may not have been trained on.

Despite the demonstrated effectiveness of these techniques, empirical research has also revealed troubling sensitivity in model behavior. Small changes in prompt wording, such as rephrasing a question or reordering examples, can produce meaningfully different outputs (Lu et al., 2022; Zhao et al., 2021). This instability raises concerns about the reliability of prompt-dependent AI outputs, because reliability requires that measurements remain consistent across irrelevant variations in the measurement procedure. If rephrasing a prompt changes the model's scoring behavior, then the prompt is functioning as a potential source of measurement error.

A significant gap in the current literature is that most prompting techniques have been evaluated on general-purpose benchmarks rather than on educationally meaningful constructs. Benchmarks such as MMLU (Hendrycks et al., 2021), GSM8K (Cobbe et al., 2021), and HumanEval (Chen et al., 2021) measure model performance on broad knowledge retrieval, mathematical reasoning, and code generation, respectively. These benchmarks are useful for comparing models but tell us little about whether a prompting technique produces outputs that are valid for a specific assessment and education purpose, such as scoring student writing against a rubric or generating items that measure a targeted learning standard. Because educational applications require alignment with specific constructs, curricular frameworks, and learner populations, the effectiveness of prompting techniques cannot be assessed purely through benchmark performance. A second gap is the absence of a design framework that connects prompting decisions to an explicit validity argument. Prompting techniques originated from the AI field are typically selected based on what works in practice rather than on a principled analysis of what the output needs to demonstrate. This means that even when a technique produces acceptable results, the reasons why it works and the conditions under which it might fail are not well documented.

**Evidence-Centered Design**

The central insight of ECD is that an assessment is an evidentiary argument (Mislevy, Almond, et al., 2003). A test score is not a direct reading of ability, but it is the conclusion of an argument that connects observable performance to inferences about what a learner knows or can do. ECD makes this argument explicit by requiring designers to specify each component of the argument at design time, before any items are written or any data are collected.

ECD organizes the assessment design process into five layers as shown in Figure 1 based on Mislevy, Almond, et al. (2003). *Domain analysis* identifies the KSAs that define the target construct, along with the situations in which those abilities are expressed and the features that distinguish different levels of proficiency. *Domain modeling* translates the results of domain analysis into formal structures, articulating the claims to be supported by the assessment and the evidence those claims require. *The conceptual assessment framework* specifies three interrelated models. *The student model* describes the variables to be measured, representing the KSAs about which inferences will be drawn. *The evidence model* specifies the observable features of student performance that count as evidence and the rules for extracting and evaluating those features. *The task model* specifies the situations, stimuli, and constraints that will elicit the target performance. *Assessment implementation* translates these models into operational tasks, scoring procedures, and reporting systems. *Assessment delivery* manages the administration and collection of evidence in practice.

**Figure 1**

*Five layers by ECD to organize the assessment design process*

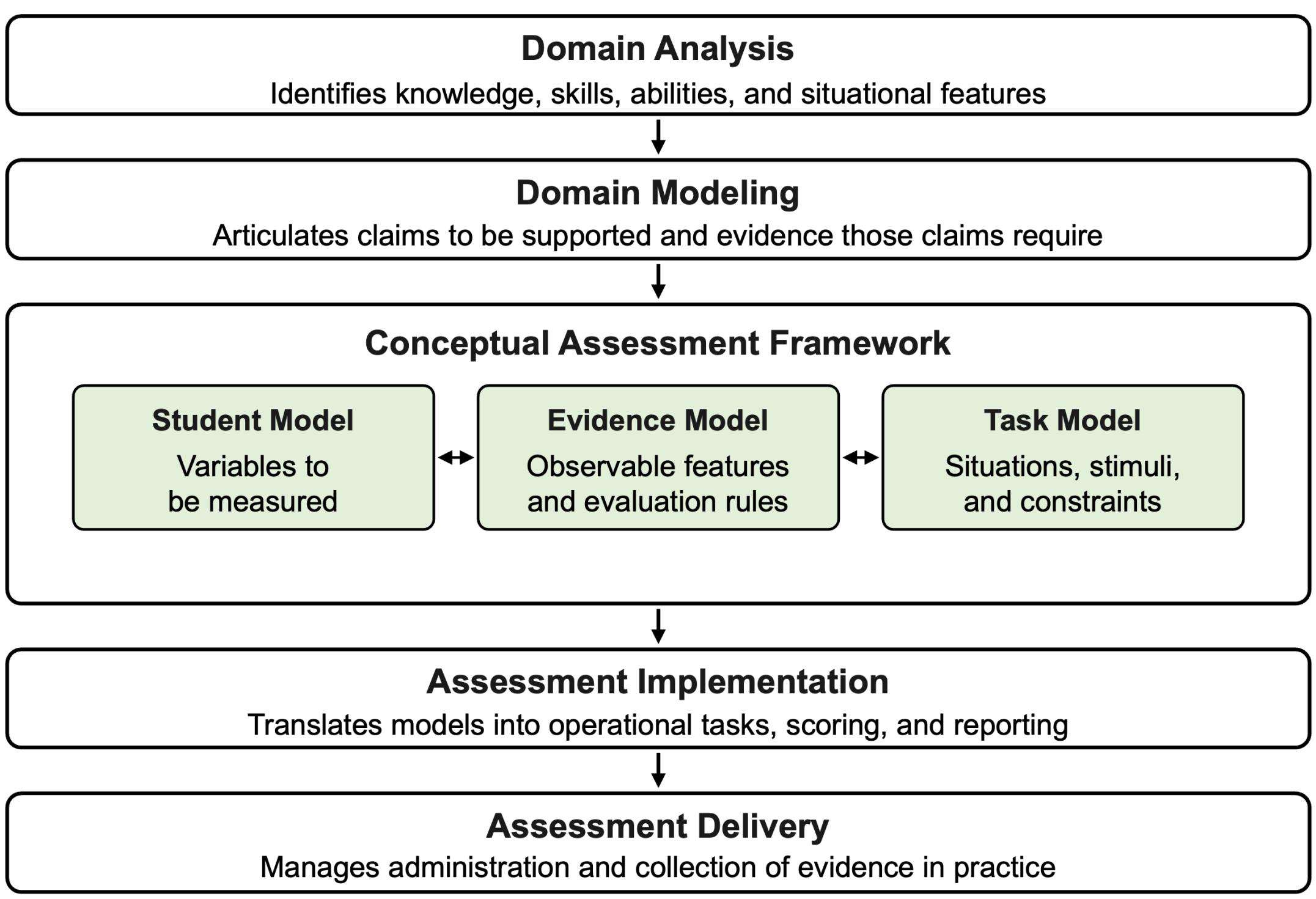

The three models at the core of the conceptual assessment framework are bound together by evidentiary argument, which is the connective reasoning that links what we want to know about the learner (i.e., the student model) with what we observe (i.e., the evidence model) and how we elicit those observations (i.e., the task model). Toulmin (2003) and Kane (2013) described similar structures for validity arguments. In Toulmin's terms, the claim states what we infer about the learner, the data are the observations from the assessment, the warrant is the reasoning rule that turns observations into evidence, the backing is the empirical and theoretical support justifying the warrant, and the rebuttal identifies alternative explanations that could weaken the inference. Kane (2013) applied this structure to validation by proposing that the intended interpretation and use of test scores be stated as an interpretation or use argument, which is a chain of inferences from observed performances to claims and decisions, and that a validity argument then evaluate the coherence of this chain and the plausibility of each inference. Messick (1989) articulated a unified conception of validity that encompassed the evidential basis for score interpretation and the consequential basis for score use. ECD operationalizes these validity frameworks by requiring designers to build the argument into the assessment from the start rather than evaluating it post hoc.

### *ECD for Prompt Engineering*

We advocate that the logic behind assessment design and prompt design are similar. Both activities require explicit articulation of what is to be inferred from the output, what evidence supports that inference, and what tasks or prompts elicit the necessary evidence. In assessment design, the designer must specify the construct before writing items, define scoring criteria before collecting responses, and establish the evidentiary argument before interpreting scores. In prompt design, the user must specify what the model should produce, what criteria the output should satisfy, and how the prompt structure elicits an output that meets those criteria. When any of these specifications is missing, the resulting output cannot be defended as appropriate and valid for its intended educational use. Further, the detailed specification assures the reproducibility of AI models' output.

Thus, prompting a LLM is like administering a performance assessment task to a LLM. The model is the respondent, and the prompt is the task that elicits a model's performance. The output is the work product that must be evaluated against pre-specified criteria. From this perspective, the prompt functions as a measurement task that must be designed with the same rigor as any other assessment task. For instance, if a rubric for human scoring would never be deployed without careful specification and pilot testing, then a prompt that governs automated scoring should not be deployed without analogous design scrutiny.

### *Applications of ECD in AI-Related Contexts*

Some recent work has begun to connect ECD with AI applications in education, though most focus on using ECD to structure AI-generated products rather than the prompts that drive them. Huff et al. (2013) demonstrated how ECD task models can guide automatic item generation in the pre-GenAI era by specifying the construct-relevant features that generated items must preserve. Their work established the principle that automated generation does not eliminate the need for principled design specifications. In fact, those specifications become the constraints that govern what the non-GenAI based generation algorithm is permitted to produce. Learning from that, recent work has applied ECD to structure LLM-based conversation assessments, using the evidence model to define what conversational moves count as evidence and the task model to specify how the LLM should elicit those moves (Roohr et al., 2024). In the domain of feedback generation, ECD has been used to design LLM-based automated feedback systems for physics problem solving, grounding the feedback in explicit evidence rules that link student errors to targeted instructional responses (Maus et al., 2025). Multi-agent systems for science assessment development have employed ECD to coordinate multiple LLM agents, with each agent responsible for a different component of the design process (Yang et al., 2026). Hao (2025) provided a broader analysis of opportunities and challenges for AI in educational assessment through the lens of ECD and measurement values, arguing that algorithmic decisions in assessment must be traceable to the same evidentiary standards that govern human decisions.

A common theme across these applications is that ECD provides the design specification that constrains AI behavior. Without ECD, an LLM that generates items has no principled basis for determining construct relevance. ECD addresses this problem by requiring that every AI-generated output be traceable to an identifiable element of the student model, the evidence model, or the task model. However, existing work has mainly applied ECD to the outputs that AI generates, and the prompt that produces those outputs has received little evidentiary scrutiny as a design artifact. This omission matters because the prompt is the primary mechanism through which users control model behavior. If the prompt is poorly designed, even a capable model can produce outputs that fail to meet the evidentiary standards that principled assessment design requires. This work contributes to closing this gap by proposing an ECD-based framework for prompt engineering.

## Proposed ECD-Based Framework for Prompt Engineering

We propose a principled framework for prompt engineering to strengthen validity argument based on ECD as articulated by Mislevy, Steinberg, et al. (2003) and extended through the validity frameworks of Messick (1989), Toulmin (2003) and Kane (2013). The core principle is that any inference drawn from an assessment including performance tasks, or in our case from an AI-generated output, must be supported by an explicit chain of reasoning that connects the observed evidence to the intended claim. This chain of reasoning constitutes the evidentiary argument. Under ECD, the evidentiary argument should not be evaluated post hoc. It is built into the design before any data is collected.

Applying this principle to prompt engineering means treating every prompt as an assessment instrument or performance task design that must satisfy a validity argument. The claim is the intended use of the AI output. For example, if a GenAI model or LLM is prompted to score student essays, the claim is that the scores reflect the construct defined by the scoring rubric. The evidence is the specific features of the model's output that support or fail to support this claim, such as whether the model identifies the correct rubric elements and applies them consistently within and across-individual student writings. The warrant is the reasoning that connects the prompt design to the observed output quality. Prompt design elements constitute facets for the intended construct the AI model is prompted to look for and evaluate on as they determine what information the model receives, what constraints it operates under, and what standards it is asked to meet. The backing includes empirical demonstrations that the prompt produces outputs consistent with the claim, such as agreement with human experts. The rebuttal identifies conditions under which the prompt might fail, such as sensitivity to rephrasing or differential performance across student subgroups.

This theoretical grounding distinguishes our framework from purely technical approaches to prompt optimization. Techniques like CoT and few-shot prompting are valuable tools, but they still focus on AI models' capabilities of reasoning and learning from the shots provided without a principled framework explicitly stating what the model should look for in performing the tasks in a reliable and valid way. They tell the user how to structure a prompt but not what the prompt should accomplish, what evidence the output should contain, or how to evaluate whether the prompt has achieved its purpose. ECD provides this missing layer by embedding each prompting decision within a validity argument that can be inspected and revised.

**Mapping ECD Components to Prompt Design**

The translation of ECD into prompt engineering requires mapping each of the three core models onto corresponding elements of prompt design. We describe each mapping in turn and explain the reasoning behind it.

The student model specifies the KSAs that the scores are intended to measure. In automated essay scoring, the student model includes the overall construct and its facets, such as organization, development of ideas, and use of conventions. A scoring prompt makes the student model explicit by defining the construct, the dimensions to be scored, and features of a response fall outside the construct. For example, a prompt for scoring the development of ideas should state whether grammatical errors count against that dimension. Without an explicit student model, the LLM has to infer what the score is about, and the score may then reflect construct-irrelevant features of the response.

The evidence model, in prompt engineering, specifies how a student's response becomes evidence about the student model. Mislevy, Steinberg et al. (2003) described two components of the evidence model. The evaluation component contains the evidence rules that identify the salient features of a work product and express them as observable variables. The measurement component then relates these observable variables to the student model variables. In a scoring prompt, the score-point descriptors of the rubric serve as the evidence rules, because they state which features of a response justify each score on each dimension. Example responses that illustrate each score point also belong to the evidence model, because they show how the rules apply to actual student writing. The prompt should therefore instruct the LLM to identify the features of the response that match a descriptor, quote that evidence, and then assign the score. The reasoning that the LLM reports makes this step visible, so reviewers can check whether each score rests on construct-relevant features. The measurement component usually lies outside the prompt, because it governs how the dimension scores from the LLM are combined or scaled to support claims about the student.

The task model describes the situation that elicits the student's performance. Mislevy, Steinberg et al. (2003) described the task model as specifying the materials presented to the student, the features of the task that affect what evidence it can provide, and the work product, which is the form in which the student's response is captured. In automated essay scoring, the LLM does not present the task to students, but it needs the task model to interpret their responses. For example, an essay that closely paraphrases a source text may meet the requirements of a task that asks students to explain the source, but it would fall short on a task that asks students to argue a position. The scoring prompt should therefore include the writing task, the source materials, the grade level, and the response requirements that students received. The task model has a larger role when the LLM generates items, because the prompt then specifies the features of the tasks that the LLM is asked to produce.

The evidentiary argument ties these three models together. In ECD-based prompt engineering, it is the explicit rationale linking the prompt design to the intended educational use of the output. It answers the question of why this particular prompt, with these particular components, should be expected to produce output that satisfies the evidence model and supports the claims specified in the user and context model. The evidentiary argument makes the prompt a defensible instrument whose design can be inspected, evaluated, and improved.

**ECD-Based Prompt Engineering Framework**

Our proposed principled framework is presented in Figure 2 to operationalize ECD for prompt engineering, using automated essay scoring as an illustration. The full prompt for this example can be found in Appendix A. Mapped to the ECD layers in Figure 1, domain analysis and domain modeling play the same roles that they play in assessment design. Prompt specification takes the place of the conceptual assessment framework, because it is the layer in which the student model, the task model, and the evidence model are specified. The output layer corresponds to assessment delivery, in which the LLM applies the prompt to each student response and returns a score. This framework involves evidentiary arguments at two levels, and separating them clarifies how ECD applies to prompting. The first argument concerns the student. It is stated inside the prompt, and the LLM builds it for every response by identifying evidence in the response and linking that evidence to a score point. This argument corresponds to the scoring inference described by Kane (2013). The second argument concerns the prompt. The designer formulates it during domain modeling and supports it with evidence from the outputs, such as agreement between LLM scores and the scores of trained human raters.

Prompt specification translates the products of domain analysis and domain modeling into the prompt. Every part of the prompt belongs to this specification, and each part has a defined evidential function. The prompt starts with scoring instructions that describe its seven parts, and each part has a defined evidential function. Part 1 specifies the student model, which has one variable, proficiency in narrative writing based on a source text. The score is a claim about the level of this proficiency that a certain grade student demonstrated in this one response, and the LLM is told not to reward or penalize features outside the trait. Part 2 specifies the task model, which consists of the source text and the writing prompt. It also names the task model variables, including the demand to retell the babysitting experience from another character's point of view. Part 3 specifies the evidence model. Rubric supplies the evidence rules, so the LLM checks the condition codes under Score 0 before matching the observed features to the descriptions for Scores 1 through 4. The measurement model is simple for this item, because the holistic score is reported directly as evidence about the student model variable. Part 4 states the evidentiary argument. The claim is the score point, the data are the observable features in the response, the warrant is the rubric, and the backing is the set of example responses in Part 5. The LLM identifies the evidence, maps it to a score point, and then states the rationale. Part 4 also includes a step that the LLM checks whether something other than the target proficiency could explain the observed evidence, such as copying, response length, or errors in conventions.

In particular, Part 5 applies the same analysis structure to the examples. In conventional few-shot prompting, an example pairs a response with its score, and the model must infer why the response earned that score. In our prompt, each anchor response is followed by an ECD analysis that records the observed evidence, the rationale that applies the rubric, and the score. Each anchor is therefore a complete evidentiary argument. Part 6 requires output schema with three keys, and each key records one element of the argument. When a response lies near a score boundary, the rationale must also explain why the adjacent score point fits less well. Part 7 contains the response to be scored. Each output is therefore an auditable record of the evidentiary argument. Across responses, the outputs supply the evidence specified in domain modeling, such as agreement with human raters.

**Figure 2**

*ECD components for prompt engineering*

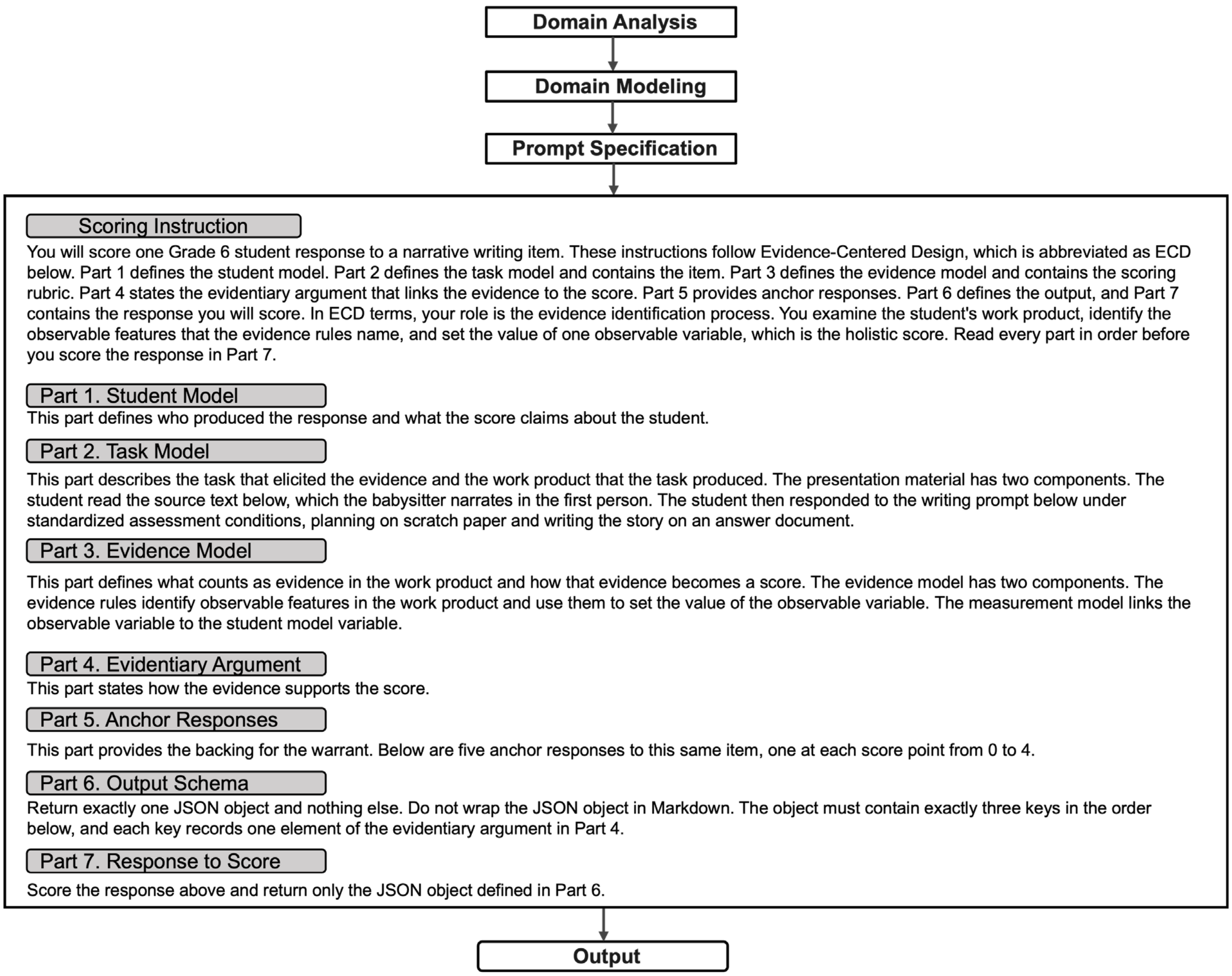

## Design Principles Derived from ECD

Our proposed ECD-based framework yields five design principles that guide prompt engineering practice. These principles are not prescriptive rules but rather standards against which prompt design decisions can be evaluated.

The principle of intentionality means that every component of a prompt should serve a stated inferential purpose. If a prompt includes a role specification, the designer should be able to explain what evidential function that role serves. If a prompt includes examples, the designer should be able to explain why those particular examples were chosen and what aspect of the intended output they calibrate. Components that cannot be linked to an evidential function are candidates for removal, because they may introduce construct-irrelevant noise.

The principle of traceability tells us that each design decision should be linked to the evidentiary argument. A reviewer examining the prompt should be able to trace each component back to a specific claim, evidence requirement, or task specification. Traceability supports accountability because it makes the reasoning behind the prompt visible and open to scrutiny. It also supports collaboration, because designers can communicate not just the prompt itself but the reasoning that produced it.

The principle of accountability indicates that outputs should be evaluated against pre-specified criteria rather than post hoc impressions. It is common practice to evaluate AI outputs by reading them and judging whether they seem reasonable. This approach is vulnerable to confirmation bias and lacks the rigor needed for educational applications. Accountability requires that the evidence model specify the evaluation criteria before the outputs are generated, so that evaluation is anchored in explicit standards rather than subjective judgment.

The principle of iteration suggests that prompt design is an empirical and revisable process. A prompt is not a static artifact. It is a hypothesis about what instructions will elicit the intended output from a particular model on a particular task. Like any empirical hypothesis, it must be tested and revised based on evidence. The iterative nature of ECD-based prompt engineering distinguishes it from one-shot prompt writing, in which the user issues a prompt, evaluates the output informally, and either accepts or rewrites it without systematic analysis.

The principle of equity holds that evidence collection must include checks for differential performance across learner subgroups. Educational assessments are expected to function equitably across subgroups of the population, and the same expectation applies to AI-generated educational outputs. An automated scoring prompt that produces accurate scores for one student group but biased scores for another is not valid for general use, regardless of its overall accuracy. The equity principle requires that prompt design explicitly incorporate fairness analyses as part of the evidence evaluation step.

## Empirical Illustration with Automated Scoring

### Research Questions

To examine the practical value of the ECD framework in guiding prompting strategy, we conducted an empirical data analysis using automated scoring as an example. Specifically, this demonstration addresses two research questions (RQ):

RQ1: How can ECD be used to guide prompt engineering frameworks for automated scoring tasks?

RQ2: Do ECD-guided prompts yield higher scoring accuracy than the SOTA prompting methods?

### Study Design

This study employs a two by three factorial design. The first factor is ECD guidance, representing whether the prompt was designed through the seven-step ECD workflow described above. The second factor is prompting strategy, indicating commonly used prompt strategies in educational research. The specific information is listed in Table 1, which yields six different prompt conditions. All prompt conditions are administered through the GPT-OSS-120B model, accessed via API with the temperature parameter set to 0.0 to control LLM’s stochastic variation when outputting results. The prompt details are attached in the Appendix, in which item stem information and scoring rubrics were embedded.

**Table 1**

*Study design with factors and levels*

| Factor | Levels |
| --- | --- |
| ECD guidance | ECD-guided prompt design |
| | No-ECD guided |
| Prompting Strategy | Zero-shot prompting |
| | Few-shot prompting |
| | Chain-of-thought |

### *Dataset and Human-Rater Benchmark*

The dataset used in this analysis consists of student responses to extended constructed-response items from an English Language Arts interim assessment administered in a southeastern U.S. state at Grades 6, 7, and 8. Each grade has one item examining students' narrative writing. Each item presents students with a reading passage and asks them to write a portion of the narrative using descriptive details. All items are scored on a five-point holistic rubric ranging from 0 to 4. The rubric evaluates students' ability to develop real or imagined experiences using narrative techniques such as dialogue, description, and pacing, along with clear event sequences, precise language, and effective integration of source material.

The full response pools contain 3,774 responses at Grade 6, 6,532 at Grade 7, and 4,643 at Grade 8. Each response was scored by two trained human raters. Across all three grades, inter-rater exact agreement rates were consistent at approximately 91.4%. Quadratic weighted kappa (QWK) coefficients were 0.96, 0.94, and 0.95 for Grade 6. 7 and 8, respectively.

From each grade-level pool, we selected a stratified sample of 500 student responses for the LLM scoring study. The sampling preserved the score distribution of the full population so that the 500 responses at each grade level reflect the same proportional representation across score points as the complete dataset. The final scores for these 500 responses serve as the benchmark against which all LLM-assigned scores are compared

### *Evaluation Metrics*

We evaluated LLM scoring quality using five metrics. Exact agreement is the proportion of responses for which the LLM-assigned score matches the human assigned score exactly. Adjacent agreement relaxes the criterion to allow scores within one point of the human assigned scores. QWK measures the agreement between LLM-assigned and human scores after adjusting for agreement expected by chance. QWK values range from negative, indicating less agreement than expected by chance, to 1.0 for perfect agreement. In operational automated scoring contexts, QWK values above .70 are generally considered acceptable, and values above .80 indicate strong agreement (Williamson et al., 2012). Spearman rank-order correlation ($\rho$) quantifies the monotonic relationship between score sets without assuming linearity. For ordinal data such as rubric scores, Spearman correlation provides a distribution-free measure of whether the model preserves the rank ordering of responses established by human raters.

***Results***

Table 2 presents the scoring quality metrics for all prompt conditions across three grades. The results show a consistent pattern in which ECD-guided prompts outperformed their non-ECD counterparts across all metrics and grade levels. We can see that, across all three prompting strategies, the ECD-guided versions produced higher accuracy and consistency values than the corresponding non-ECD baselines. Additionally, among the non-ECD conditions, few-shot prompting achieved the strongest performance, zero-shot prompting produced moderate agreement, and CoT prompting without ECD guidance yielded the weakest results.

**Table 2**

*Scoring quality metrics by grade and prompt condition*

| Grade | Prompting Strategy | ECD | Exact Agreement | Adjacent Agreement | QWK | Spearman $\rho$ |
|---|---|---|---|---|---|---|
| 6 | Zero-Shot | No | 0.422 | 0.854 | 0.626 | 0.744 |
| | Few-Shot | No | 0.432 | 0.912 | 0.700 | 0.759 |
| | Chain-of-Thought | No | 0.376 | 0.776 | 0.500 | 0.715 |
| | Zero-Shot | Yes | 0.492 | 0.866 | 0.679 | 0.769 |
| | Few-Shot | Yes | **0.604** | **0.934** | **0.814** | **0.832** |
| | Chain-of-Thought | Yes | 0.464 | 0.856 | 0.654 | 0.761 |
| 7 | Zero-Shot | No | 0.452 | 0.856 | 0.635 | 0.730 |
| | Few-Shot | No | 0.486 | 0.878 | 0.691 | 0.750 |
| | Chain-of-Thought | No | 0.420 | 0.776 | 0.524 | 0.607 |
| | Zero-Shot | Yes | 0.536 | 0.890 | 0.727 | 0.769 |
| | Few-Shot | Yes | **0.708** | **0.932** | **0.845** | **0.859** |
| | Chain-of-Thought | Yes | 0.480 | 0.824 | 0.622 | 0.680 |
| 8 | Zero-Shot | No | 0.504 | 0.846 | 0.639 | 0.716 |
| | Few-Shot | No | 0.530 | 0.872 | 0.710 | 0.762 |
| | Chain-of-Thought | No | 0.446 | 0.782 | 0.547 | 0.672 |
| | Zero-Shot | Yes | 0.576 | 0.868 | 0.706 | 0.748 |
| | Few-Shot | Yes | **0.786** | **0.944** | **0.886** | **0.889** |
| | Chain-of-Thought | Yes | 0.584 | 0.832 | 0.671 | 0.718 |

In general, the addition of ECD guidance improved all three prompting strategies, but the magnitude of improvement varied. The largest gains appeared in the ECD-few-shot condition, which achieved QWK values of 0.814 at Grade 6, 0.845 at Grade 7, and 0.886 at Grade 8, respectively. Exact agreement for ECD-few-shot ranged from 0.604 to 0.786, and adjacent agreement exceeded 0.930 at all grades. Spearman correlations followed the same ordering as QWK across all conditions. ECD-few-shot produced the highest correlations, with Spearman $\rho$ values of 0.832, 0.859, and 0.889 across the three grades.

**Figure 3**

*Score distributions of human benchmark scores and LLM-assigned scores across six prompt conditions*

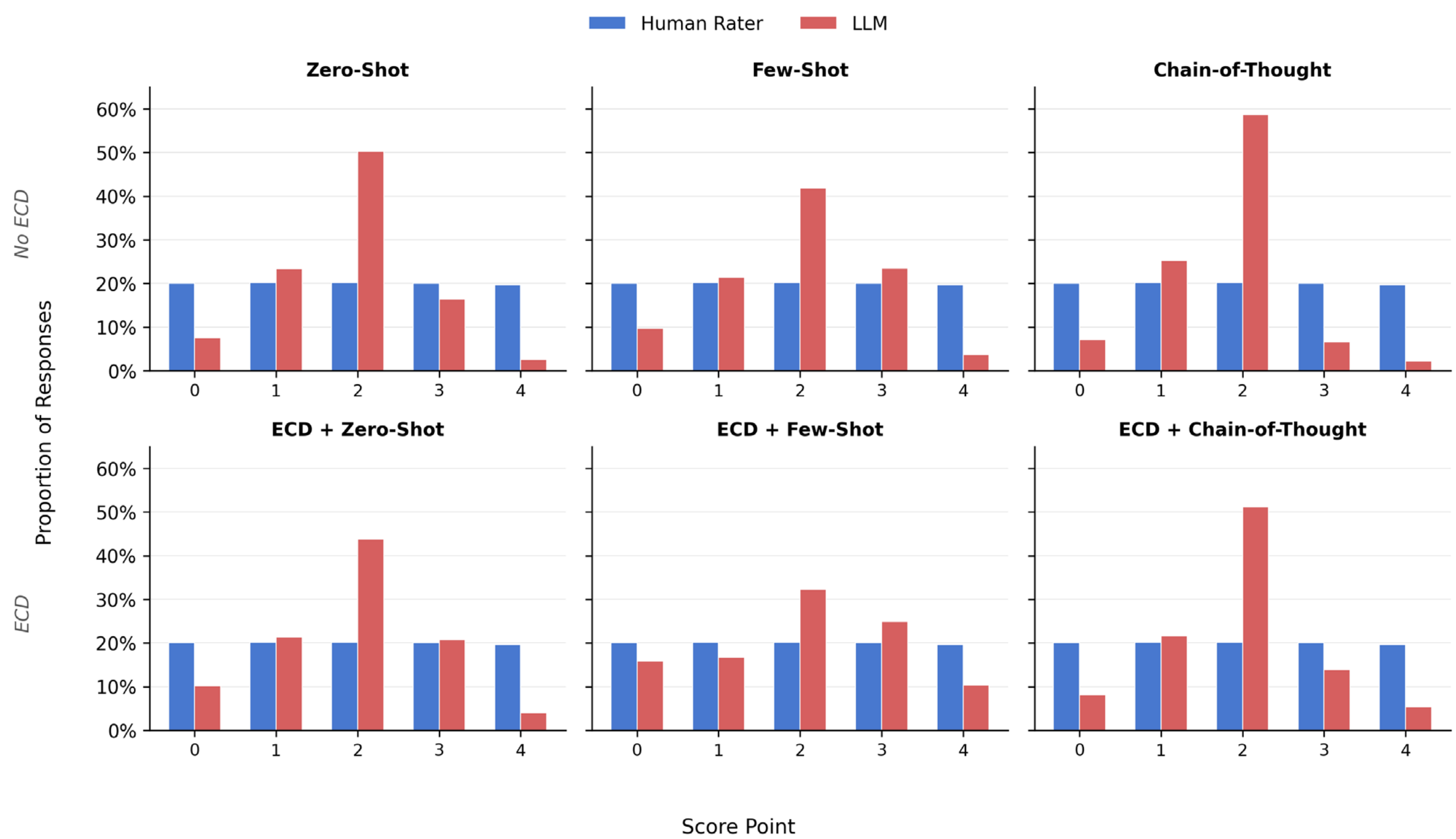

To examine the distributional properties of LLM-assigned scores, we compared the score distributions produced by each condition against the human benchmark distribution. Figure 3 displays these comparisons collapsed across grades. All six conditions over-assigned scores of 2 relative to the human benchmark, which reflects a central tendency bias in LLM scoring in which models default to middle-range scores at the expense of extreme score points. The severity of this bias varied substantially across conditions. The non-ECD with chain-of-thought condition exhibited the most extreme central tendency, assigning approximately 59% of responses a score of 2 compared to the human benchmark rate of 20%. The ECD-few-shot condition produced the most balanced score distribution among all conditions, with the smallest discrepancies between human and LLM proportions at each score point. This indicates that ECD-few-shot reproduced the overall shape of the score distribution more faithfully than other conditions.

Figure 4 presents the confusion matrices for the six conditions collapsed across grades. Each cell shows the count and its percentage of the 1500 responses in that condition, so the diagonal cells sum to the exact agreement rate. In the three non-ECD conditions, more responses with human scores of 3 or 4 were scored as 2 than were scored correctly at those two levels, and more responses with a human score of 0 were scored as 1 than as 0. This pattern indicates that the model without ECD guidance lacked the specificity to distinguish score levels at the extreme scores. The ECD-few-shot condition showed the highest diagonal consistency. Of the responses scored correctly under ECD-few-shot, more score points of 0 and 4 were consistent with human scorers, which suggests that ECD guidance improves scoring accuracy most in boundary cases. Grade-specific confusion matrices are presented in Tables B1 to B3 in Appendix B.

**Figure 4**

*Confusion matrices for six prompt conditions collapsed across grades*

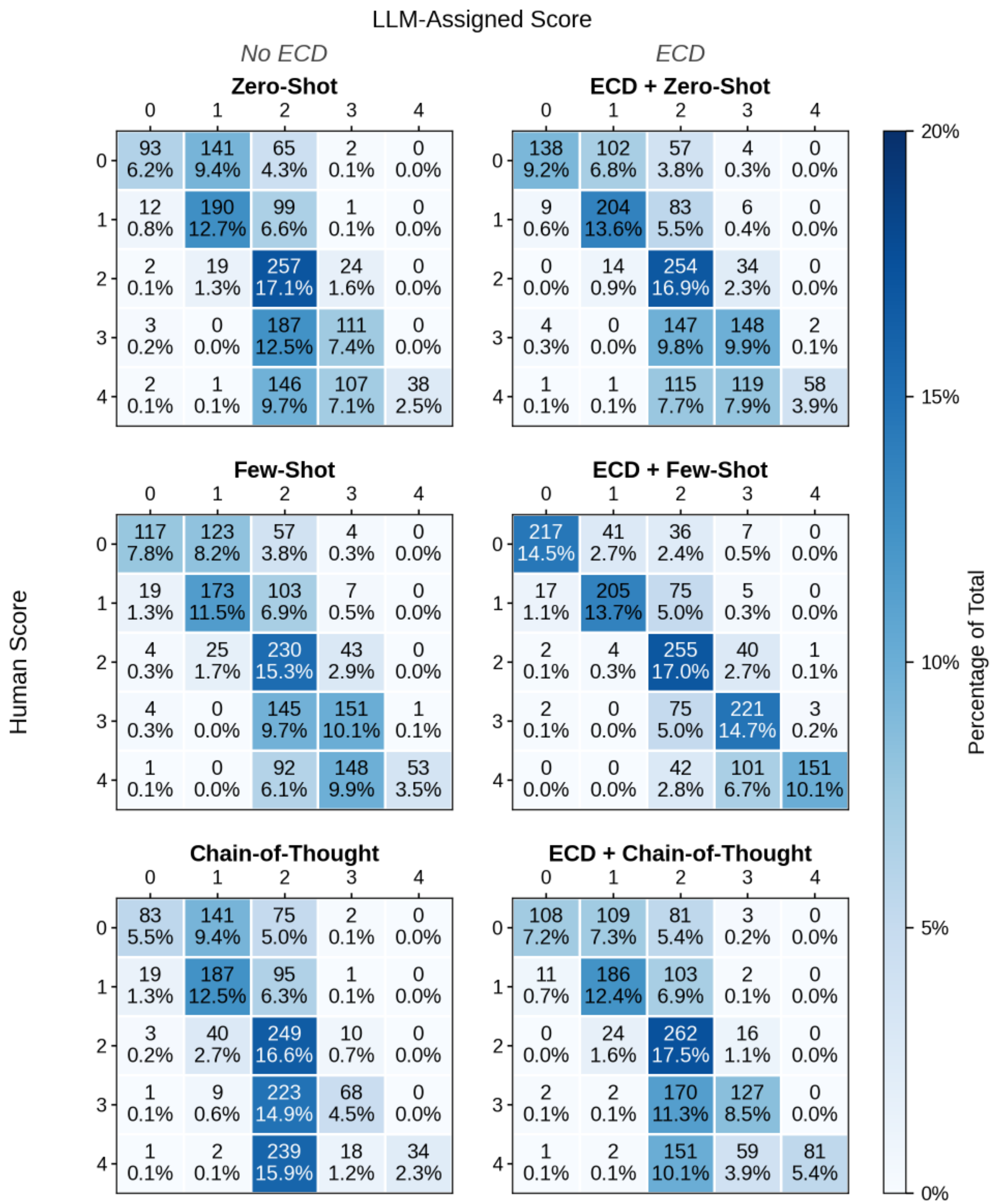

## Discussion

The framework and empirical evidence presented in this paper reposition prompt engineering as an evidentiary design activity. The empirical results confirm that ECD consistently improved scoring quality across all prompting strategies and grade levels. Chain-of-thought prompting without ECD guidance produced the weakest results of any condition, suggesting that explicit reasoning steps could be less useful when the model lacks a well-specified evidential standard to reason toward. Few-shot prompting benefited most from ECD guidance because the framework provides principled criteria for selecting exemplars rather than choosing them by convenience.

Current practice treats prompt engineering as conversational strategy of LLMs without design principles for high stakes uses in education. Users learn through experimentation, community tips, and pattern matching from successful examples. This approach can produce acceptable results in low-stakes contexts, but it provides no principled basis for trustworthy outputs in educational applications where accuracy, fairness, and alignment with learning objectives matter. By embedding prompt design built upon the ECD framework, we provide a principled approach that connects every design element to an explicit validity argument. The prompt is no longer a conversational instruction whose effectiveness is judged by individual users. Rather, it is designed with validity evidence whose adequacy can be evaluated against pre-specified criteria.

A contribution of this work is the conceptual connections it builds between educational measurement and human-AI interaction. Educational measurement has developed sophisticated frameworks for ensuring that inferences drawn from observed performance are valid, reliable, and fair. Human-AI interaction research has focused on how users communicate with AI systems to achieve desired outcomes. These two fields have developed largely in parallel, but the rise of AI in educational settings creates an urgent need for integration. When an educator prompts a LLM to score student work, that interaction simultaneously involves measurement concerns (e.g., is the score valid?) and interaction concerns (e.g., did the prompt effectively communicate the task?). ECD-based prompt engineering addresses both concerns within a single principled design framework.

The ECD framework implies that all responsible users of AI in education must be designers of their interactions with AI systems. This is a stronger claim than the common recommendation that users should learn prompt engineering techniques. Knowing techniques is necessary but not sufficient. Users must also possess adequate substantive content knowledge to know what they want the AI to produce, what evidence would indicate that the output is adequate, and how their prompt design supports the connection between the two. In other words, responsible AI use requires the kind of design thinking that ECD provides.

This implication extends naturally to AI literacy curricula and teacher professional development. Current AI literacy efforts typically focus on understanding what AI can and cannot do, recognizing AI-generated content, and learning basic prompting techniques. The ECD framework suggests that AI literacy should also include the ability to articulate the intended use of an AI output, specify the evidence criteria the output should satisfy, evaluate outputs against those criteria, and document design decisions for transparency. These competencies align with assessment literacy, which is already a recognized component of educator preparation. Integrating ECD-based prompting into teacher professional development programs would leverage existing assessment literacy skills and extend them to the application of AI to domain-specific context.

The framework also implies that all AI users need to be critical evaluators of AI outputs. The evidentiary argument structure requires that outputs be evaluated against pre-specified criteria, not accepted at face value. This critical stance is important because LLMs produce outputs that are fluent and confident regardless of their accuracy. Without a framework for evaluation, users may accept outputs that sound authoritative but are factually wrong, misaligned with the target construct, or biased in ways that disadvantage particular students.

The design principles we have articulated also have implications for the developers of educational AI tools. If principled prompt design improves the validity of AI outputs, educational AI interfaces should support users in applying these principles. This might include structured prompt templates that guide users through the domain analysis and evidence specification steps, built-in evaluation tools that compute agreement metrics against benchmark responses, and documentation features that allow users to record the evidentiary argument behind their prompts. Supporting transparent documentation of prompts and their evidentiary arguments would enable institutional review, collaborative refinement, and accumulation of knowledge about which prompt designs work for educational purposes.

Several limitations of the present work warrant discussion. The empirical findings are bounded by the specific tasks and models studied. Automated scoring is a relatively well-structured task with explicit rubrics and available benchmarks. Whether the ECD advantage generalizes to less structured tasks such as open-ended feedback generation or lesson planning remains an empirical question. The models used in the study represent a snapshot of one LLM, and GenAI models' capability updates may shift the relative effectiveness of different prompting approaches. Because model providers frequently update their systems, a prompt optimized for one model version may perform differently on a subsequent version, which reinforces the importance of the monitoring step in the ECD workflow.

The cost and expertise required to apply ECD in everyday classroom settings presents a practical barrier. The seven-step workflow demands time, assessment knowledge, and iterative testing that individual teachers may not have. This limitation is real, and we do not minimize it. However, it is analogous to the challenge of developing high-quality assessment tasks, which also requires expertise and iterative refinement. Practical solutions include the development of pre-designed ECD prompt templates for common educational tasks, institutional support for prompt design as part of instructional design workflows, and shared repositories of validated prompts with their evidentiary documentation.

## Conclusion

Generative AI has been increasingly applied in assessment and education, and the quality of its output depends substantially on how users construct the prompts that elicit model behavior. Current prompt engineering practices lack the principled design structure needed to ensure that AI outputs are valid, reliable, and fair for their intended educational uses. The framework presented in this paper addresses this gap by translating ECD into the domain of prompt engineering. We have shown that the three core models of ECD, the student model, the evidence model, and the task model, map coherently onto the components of prompt design. The student model is a user and context model that specifies who the learner is and what the output should accomplish. The evidence model is explicit output specifications that define what the response should contain and how it should be evaluated. The task model is the prompt structure itself, with each component serving an identifiable evidential function. The evidentiary argument ties these models together into a rationale that can be inspected, challenged, and refined.

Responsible educational use of generative AI requires principled design of the prompts themselves. A model is only as good as the instructions it receives, and those instructions are only as good as the design behind them. ECD provides the design thinking that prompt engineering currently lacks. It connects established measurement science with an emerging practice, offering a transferable framework that applies wherever educators, researchers, or developers use AI to support teaching, learning, or assessment.

Several directions for future research emerge from this work. For example, developing ECD templates for non-expert users would address the accessibility barrier and enable broader adoption of principled prompt design. Such templates would pre-populate the ECD structure for common educational tasks, allowing users to customize the content while maintaining the evidentiary framework. Research on tooling for automated ECD-based prompt authoring and validation would further reduce the burden on individual users by providing software support for the design workflow. Additionally, longitudinal studies of ECD-prompted AI in authentic classroom settings would provide evidence about the framework's practical impact over time. Such studies could examine whether teachers who use ECD-based prompts produce more consistently valid AI outputs, whether students who receive ECD-designed AI feedback show different learning trajectories, and whether the framework influences teachers' broader assessment literacy. Finally, extending ECD-based prompting to multimodal and agentic AI systems represents a forward-looking research frontier. As AI systems increasingly process and generate images, audio, and interactive content, the ECD framework must accommodate evidence models for multimodal outputs. Similarly, as AI agents take on multi-step educational tasks with greater autonomy, ECD must address how to structure the evidentiary argument for sequences of AI decisions rather than single outputs.

# Appendix A

## A.1 Condition A: Zero Shot Prompt

Score the student response on a 0–4 scale using the item and holistic rubric provided below.
---
## Item Information — Grade 6

### Source Text: The Entrepreneurial Problem Solver

As a young enterprising entrepreneur, I quit pestering my parents for money eight months ago. Since March, I have been helping with assorted jobs around the neighborhood. I started in the spring by helping Mr. Arbuckle seed his garden. Soon, I was mowing lawns, and once autumn came, I was literally raking it in at $20 a lawn. But November arrived, and my cash flow evaporated. There are no lawns to mow, no leaves to rake in November, so I decided to seek out new employment opportunities.

That's when I discovered the perfect winter job: babysitting. Babysitters get paid to play with children, eat dinner, and watch television. What job would be better than that?

My first experience was working for the Wurthers. The job was problematic because I'm barely six months older than Jason, the oldest of two brothers, and he understandably resented the fact that his parents paid me to chaperone him and his little brother Robert. However, after 30 minutes of constant rebellion, I had the brilliant idea of sharing my duties with Jason. I presented the idea of our sharing the job and promised to compensate him by paying him half my salary. He loved that, and before I knew it, Jason was working harder than I was. When I explained my solution to Dad on the way home, he laughed and called it sub-contracting. While it cost me $20, it certainly made my life easier.

After my experience at the Wurthers, I believed I could handle anything. My mother even joked that I had earned my full credentials as a babysitter. However, it was not until babysitting Connor Doverson I learned my true capabilities.

I arrived at the Doversons' house five minutes early, and they introduced me to their son Connor and provided a tour of their house. Connor is three years younger than I, so I had assumed this job was going to be quite effortless. We would play games or watch a movie. What kind of trouble could he possibly make?

After dinner we watched a cartoon and played a card game when Connor asked if we could play hide and seek. Hide and seek is an ideal babysitting activity. Time seems to evaporate when you are hunting, so I volunteered to count first. However, Connor wanted me to hide, so I capitulated. I hid behind the door in his bedroom, so he could easily find me. When babysitting, the point of hide and seek is not to win, but to use as much time as possible; it is easier to delay the game when you are seeking than when you are hiding, so I always do a bad job of hiding. Besides, children love it when they find you.

Five minutes later, he was beaming with pride and boasting about how easily he found me, and I was counting to 20. Of course, I cheated. My job was to keep an eye on the kid at all times, so I watched as Connor slipped into a crevice between the washing machine and clothes dryer. It was a good thing I was peeking because I would never have found him there.

I began by intentionally searching in the wrong direction. I walked all over the house—even the garage where I made noise by pretending to knock a metal broom onto a tool box. I heard him giggling in the laundry room. After 15 minutes I announced my defeat and gave up. That's when the screams commenced.

I rushed to the laundry room in a panic. "What's wrong, what's wrong?" I shouted.

"I, I'm stuck!" Connor replied, tears streaming from his eyes.

Connor had wedged himself in the small space between the washing machine and the dryer. I now understood the gravity of the situation.

"Calm down," I said, as much to myself as to him. "I can fix this."

"I can't," he wailed. "I'm stu-stu-stuck."

"Connor, Connor," I heard myself saying in a soothing tone that I barely recognized as mine. "I can help you, but I need for you to calm down." This was only half true. I desperately needed him to relax, but I had no idea how to extract him from this predicament.

Gradually, Connor's screams fell to sniffles, and he eventually shifted his hand up to wipe his tears when I noticed that the dryer rocked.

"Connor," I said, "Don't move. I'm going to get something that will get you out."

I bolted to the garage and got the broom and the tool box. Soon, I returned and placed the tool box in front of the dryer and slid the broom under the appliance. I had made a lever, and slowly used it to tilt the dryer up and away from Connor. Without saying anything, Connor slid out from his hiding place and jumped into my arms. While I did not look forward to telling Connor's parents about this, I knew I was a problem-solver ready for anything the babysitting world could throw at me.

### Extended Writing Question

"The Entrepreneurial Problem Solver" is written from the point of view of the babysitter. The story would have been very different in tone and action had it been written by one of the children being babysat.

Choose another character in the story and rewrite the babysitting experience from that character's point of view. Be sure to include details that show how the story changes when experienced from another point of view.

Use the scratch paper provided to plan your writing. Then, write your story on the lines provided on your answer document.

## Rubric Information — Grade 6

### Four-Point Holistic Rubric

**Genre:** Narrative

**Writing Trait:** This trait examines the writer’s ability to effectively develop real or imagined experiences or events using effective techniques, descriptive details, and clear event sequences based on a text that has been read.

### Score 4
The student’s response is a well-developed narrative that fully develops a real or imagined experience based on text as a stimulus.
- Effectively establishes a situation and introduces a narrator and/or characters
- Organizes an event sequence that unfolds naturally
- Effectively uses narrative techniques, such as dialogue, description, and pacing, to develop rich, interesting experiences, events, and/or characters
- Uses a variety of words and phrases consistently to convey the sequence of events and signal shifts from one time frame or setting to another
- Uses precise words, phrases, and sensory language consistently to convey experiences and events
- Provides a conclusion that follows from the narrated experiences or events
- Integrates ideas and details from source material effectively
- Has very few or no errors in usage and/or conventions that interfere with meaning*

### Score 3
The student’s response is a complete narrative that develops a real or imagined experience based on text as a stimulus.
- Establishes a situation and introduces one or more characters
- Organizes events in a clear, logical order
- Uses narrative techniques, such as dialogue, description, and pacing, to develop experiences, events, and/or characters
- Uses words and/or phrases to indicate sequence of events and signal shifts from one time frame or setting to another
- Uses words, phrases, and details to convey experiences and events
- Provides an appropriate conclusion
- Integrates some ideas and/or details from source material
- Has a few minor errors in usage and/or conventions that interfere with meaning*

### Score 2
The student’s response is an incomplete or oversimplified narrative based on text as a stimulus.
- Introduces a vague situation and at least one character
- Organizes events in a sequence but with some gaps or ambiguity

- Attempts to use a narrative technique, such as dialogue, description, and pacing, to develop experiences, events, and/or characters
- Uses occasional signal words inconsistently to indicate sequence of events and signal shifts from one time frame or setting to another
- Uses some words or phrases inconsistently to convey experiences and events
- Provides a weak or ambiguous conclusion
- Attempts to integrate ideas or details from source material
- Has frequent errors in usage and conventions that sometimes interfere with meaning*

### Score 1
The student's response provides evidence of an attempt to write a narrative based on text as a stimulus.
- Response is a summary of the story
- Provides a weak or minimal introduction of a situation or a character
- May be too brief to demonstrate a complete sequence of events
- Shows little or no attempt to use dialogue, description, and pacing to develop experiences, events, and/or characters
- Uses words that are inappropriate, overly simple, or unclear
- Provides few, if any, words that convey experiences, or events, or signal shifts from one time frame or setting to another
- Provides a minimal or no conclusion
- May use few, if any, ideas or details from source material
- Has frequent major errors in usage and conventions that interfere with meaning*

### Score 0
The student's response is flawed for various reasons and will receive a condition code:
- Code A: Blank
- Code B: Copied
- Code C: Too Limited to Score/Illegible/Incomprehensible
- Code D: Non-English/Foreign Language
- Code E: Off Topic/Off Task/Offensive

*Students are responsible for language conventions learned in their current grade as well as in prior grades. Refer to the language skills for each grade to determine the grade-level expectations for grammar, syntax, capitalization, punctuation, and spelling. Also refer to the "Language Progressive Skills, by Grade" chart in Appendix A for those standards that need continued attention beyond the grade in which they were introduced.

**A.2 Condition B: Few Shot Prompt**

You need to score the student response using the rubric provided. The scoring materials are provided below.

In addition, below are 5 examples of scored responses to this same item. For each example, the score shown is the score that two independent raters agreed on.

### Example 1

**ID:** 5293

**Response:**

<student_response>
as a young enterprising entrepreneur,i quit pestering my parents for money eight moths ago since march i have been helping with assorted jobs around the naborhood i stared in the spring by helping mr.arbuckle seed his garden .soon i was mowing laws and once autumn came i was to decied to seek out new employment opportunnities.
</student_response>

**Score:** 0

### Example 2

**ID:** 5550

**Response:**

<student_response>
We were playing hiding go seek with the babysitter. I was seeking first I found her in five minutes. She was hiding behind my door. It was my turn to hide so I hide in a crevice behind the dryer. She gave up after 15 minutes. I was going to get out until I got stuck. She calmed me down and got a toolbox and a broom. She made a lever out of the toolbox and broom. She used the lever to pull the dryer up so I can slip out.
</student_response>

**Score:** 1

### Example 3

**ID:** 5610

**Response:**

<student_response>
Hi,my name is Conner,and I'm going to tell you about the time I got stuck in between the washer and dryer with my babysitter! It all started with a simple card game or two. We finished playing cards and decided to play hide and seek. I counted first. Don't tell him I said this,but he didn't hide very well. I mean he could have went somewhere else besides behind the door! But anyways,it was my turn,and I decided to hide where nobody would ever find me. In between the washer and dryer!He searched the whole entire house,even the garage! Finally, he gave up and when I tried to go get him,I couldn't move! I was stuck! I couldn't help myself as I started to scream and cry. Then my superhero babysitter came to the rescue! He told me to calm down,and I finally did. He was able to get me out by lifting the dryer away from me with a mop. I jumped into his arms,thanked him for saving me,and gave him a big bear hug.
</student_response>

**Score:** 2

### Example 4

**ID:** 5279

**Response:**

<student_response>
"I love playing hide and seek with my babysitter " I thought to myself as I peeked into the laundry room trying to find her . "Nope not in there " I said aloud . I took of up the stairs to my room to try and find her up there " Boo!" I yelled as I flung the door open to find her there . "AH" she yelled in surprise , she was so sad I found her , I'm the best at this game . "your turn to count !" I yelled to her from the bottom of the stairs . "I'm so fast , she is a slow poke " I thought to myself as she reached the last step . I ran to go hide as she put her face up to the wall , I stumbled into the laundry room , forgot their was a step there . I squeezed myself into the middle of the washer and dryer ," this was a perfect hiding spot " I thought to myself . I waited and waited for her to find me , it took her a million years to finally give up , I knew I was to go for her to find me , I tried to push my self out of the middle of the machines when all of a sudden I relies ed that I could not do that , I pushed harder again , nothing mover other than the dryer . " help !" I screamed to the babysitter " I'm st stu stuuucckk !" I yelled out loud . The baby sitter ran to the garage and came back quickly , she wedged the broom somehow to get it to lift up so I could get out , when I got out I have her a huge hug , she basically saved my life . I love my new babysitter ..............
</student_response>

**Score:** 3

### Example 5

**ID:** 5194

**Response:**

<student_response>
"Connor, sweetie! Come meet the babysitter!" my mom's voice boomed up the stairs to my room. I hope this guy or girl is like those cool ones from "Adventures in babysitting" or something like that. They get chased by bad guys, lose the mom's car, and almost get arrested! This of course won't happen, but I hope some action does.  "Coming Mommy!" I reply while walking down the stairs. When I reach the last step I see it is guy who looks to be about 3 years older than me.  "Hey kid!" he says to me. I think he will be alright."Hi," I say shyly. "Okay, honey. We're just going to be gone for a few hours then we will come back." My mom said while kissing my forehead.  "Bye mommy! Bye Daddy!" I call to my parents on their way out. Now I turn to the babysitter, whose name I do not know and I probably never will. "Are you ready to eat?" he asks. I nod in reply. After we eat pizza mom ordered for us, we watch some "Tom and Jerry". After we finish that episode, "We should play hide 'n seek!" I announce. "What a great idea! Should I count first?" my babysitter agrees. "No I think you should hide first." I challenge. "Okay," he surrenders, "but I'm very good at hiding." After about 3 seconds I find him. He needs to work on his hiding skills. "My turn! My turn!" I say while bouncing excitedly. He counts while I hide in my favorite spot, between the washer and dryer. I thought I saw him peeking, but that option closed when he went in the completely wrong direction! It was so funny I couldn't help my giggles. After what seems like an hour he gives up. Instead of coming out laughing, I stay put and burst into a symphony of tears and screams. He comes running and says with worry, "What's wrong? What's wrong?" "I'm, I think I, I'm stuck." I sputter through tears. He tells me I have to calm down before he can fix it. So, slowly, but surely, I stop crying. "I have an idea stay right here." he says. Like I'm going anywhere. He comes back a couple minutes later with some stuff. I have no idea what is, but anything to get me out of here. He builds some kind of pulley system or maybe it's called a lever? I really don't know, but as soon as there is enough room I silently slip out and hug him. I guess I really did get that adventure.
</student_response>

**Score:** 4
---
(Below copy the Source Text, Extended Writing Question, and Four-Point Holistic Rubric from Prompt A)

### A.3 Condition C: Chain of Thought Prompt

You are an expert writing assessor and you will use the rubric to score students' narrative writing. Before assigning a score, reason through the essay's strengths and weaknesses as follows,
Step 1: Identify the narrative and evaluate its clarity.
Step 2: Evaluate how the narrative is developed.
Step 3: Evaluate organization and coherence.
Step 4: Evaluate language use and mechanics.
Step 5: Assign a holistic score (0-4) based on your analysis.
---
(Below copy the Source Text, Extended Writing Question, and Four-Point Holistic Rubric from Prompt A)

### A.4 Condition D: ECD with Zero Shot Prompt

SCORING INSTRUCTIONS

You will score one Grade 6 student response to a narrative writing item. These instructions follow Evidence-Centered Design, which is abbreviated as ECD below. Part 1 defines the student model. Part 2 defines the task model and contains the item. Part 3 defines the evidence model and contains the scoring rubric. Part 4 states the evidentiary argument that links the evidence to the score. Part 5 defines the output, and Part 6 contains the response you will score.

In ECD terms, your role is the evidence identification process. You examine the student's work product, identify the observable features that the evidence rules name, and set the value of one observable variable, which is the holistic score. Read every part in order before you score the response in Part 6.

PART 1. STUDENT MODEL

This part defines who produced the response and what the score claims about the student.

The writers are Grade 6 students in the United States who responded to an English Language Arts extended constructed response item. The student model has one variable, which is proficiency in narrative writing based on a source text. The writing trait for this item defines this proficiency as the writer's ability to effectively develop real or imagined experiences or events using effective techniques, descriptive details, and clear event sequences based on a text that has been read. The holistic rubric in Part 3 describes levels of this proficiency through the situation and characters the writer establishes, the organization of the event sequence, the use of narrative techniques, the words that signal sequence and shifts in time or setting, precise and sensory language, the conclusion, the integration of source material, and errors in usage and conventions.

The score you assign is a claim about the level of this proficiency that the student demonstrated in this one response to this one item. Judge the response against the performance expected of a Grade 6 student as the rubric defines it. Students are responsible for language conventions learned in Grade 6 as well as in prior grades. The rubric footnote refers to a chart in Appendix A that is not included in this prompt, so do not guess its contents. Judge conventions by the Grade 6 expectations that the rubric descriptions illustrate. Do not apply expectations from a higher or lower grade, and do not reward or penalize features that lie outside the trait.

PART 2. TASK MODEL

This part describes the task that elicited the evidence and the work product that the task produced.

The presentation material has two components. The student read the source text below, which the babysitter narrates in the first person. The student then responded to the writing prompt below under standardized assessment conditions, planning on scratch paper and writing the story on an answer document.

<source_text>
(Below copy the Source Text from Prompt A)
</source_text>

<writing_question>
(Below copy the Extended Writing Question from Prompt A)

</writing_question>

The task model variables for this item are the narrative genre, a source text that serves as the stimulus, Grade 6 as the target grade, and the demand to retell the babysitting experience from the point of view of a character other than the babysitter. The work product is the student's written story, which appears inside the <response_to_score> tags in Part 6.

Interpret the response only as an answer to this task. Features that the task requires of every student, such as narrating from another character's point of view and drawing on the events of the source text, are expected. Give credit for these features only to the extent that the rubric in Part 3 describes them, and do not treat them as merit beyond the rubric.

PART 3. EVIDENCE MODEL

This part defines what counts as evidence in the work product and how that evidence becomes a score. The evidence model has two components. The evidence rules identify observable features in the work product and use them to set the value of the observable variable. The measurement model links the observable variable to the student model variable.

Evidence rules. The holistic rubric below supplies the evidence rules for this item. Each score point names the observable features that count as evidence for that score. Begin by checking the conditions listed under Score 0. A response that is blank, copied, too limited to score, illegible, incomprehensible, written in a language other than English, off topic, off task, or offensive receives a score of 0 with the matching condition code. If no condition applies, identify in the response the observable features that the rubric names for Scores 1 through 4. Then assign the score point whose description as a whole best matches that evidence. Because the rubric is holistic, a response does not need to show every feature listed under a score point to receive that score. Base the score only on the features that the rubric names, and do not apply any rule that the rubric does not state.

Measurement model. The holistic score from 0 to 4 is the only observable variable for this item. The score is reported directly as the evidence about the student model variable, so no further combination or adjustment takes place within this task.

<rubric>
(Below copy the Four-Point Holistic Rubric from Prompt A)
</rubric>
PART 4. EVIDENTIARY ARGUMENT

This part states how the evidence supports the score. The claim is the score point, which describes the level of narrative writing proficiency that the student showed on this item. The data are the observable features you identify in the response. The warrant is the rubric in Part 3, which states which pattern of features supports each score point. Build the argument in this order. Identify the evidence in the response, map the evidence to a score point with the evidence rules, and then state the rationale that links the evidence to the chosen score point.

Before you finalize the score, check whether something other than the proficiency defined in Part 1 could explain the evidence you observed. Copying is one such explanation. A response that reproduces the source text without building a narrative of its own is copied and receives Code B. A response that places source details or source dialogue inside a narrative the student built is integrating source material, and the rubric credits that integration. Length is another possible explanation. Length is not a rubric criterion, so a longer response earns a higher score only when the added text supplies more of the evidence that the rubric describes. Errors in usage and conventions need the same care. The rubric treats errors as one feature among several in each score description, so judge them against Grade 6 expectations and by the frequency and interference with meaning that each score point describes, and weigh them together with the other features.

PART 5. OUTPUT SCHEMA

Return exactly one JSON object and nothing else. Do not wrap the JSON object in Markdown. The object must contain exactly three keys in the order below, and each key records one element of the evidentiary argument in Part 4.

The key "score_evidence_observed" records the data. Write a brief description of the observable features of the response that the rubric names, including the absence of a feature when that absence matters for the score. Describe only what can be observed in the response. When you quote short phrases from the response, use single quotation marks so that the JSON stays valid.

The key "score_point_rationale" records the warrant as it applies to this response. Write a brief explanation of how the observed evidence maps to the selected score point under the evidence rules. When the response lies near the boundary of an adjacent score point, state why the adjacent point fits less well. When the score is 0, name the condition code that applies.

The key "score" records the claim. Its value must be one integer from 0 to 4.

Do not invent criteria that are not included in the rubric, and do not include any text outside the JSON object.

PART 6. RESPONSE TO SCORE

<response_to_score>
[[STUDENT_RESPONSE]]
</response_to_score>

Score the response above and return only the JSON object defined in Part 5.

**A.5 Condition E: ECD with Few Shot Prompt**

SCORING INSTRUCTIONS

You will score one Grade 6 student response to a narrative writing item. These instructions follow Evidence-Centered Design, which is abbreviated as ECD below. Part 1 defines the student model. Part 2 defines the task model and contains the item. Part 3 defines the evidence model and contains the scoring rubric. Part 4 states the evidentiary argument that links the evidence to the score. Part 5 provides anchor responses. Part 6 defines the output, and Part 7 contains the response you will score.

In ECD terms, your role is the evidence identification process. You examine the student's work product, identify the observable features that the evidence rules name, and set the value of one observable variable, which is the holistic score. Read every part in order before you score the response in Part 7.

PART 1. STUDENT MODEL

This part defines who produced the response and what the score claims about the student.

The writers are Grade 6 students in the United States who responded to an English Language Arts extended constructed response item. The student model has one variable, which is proficiency in narrative writing based on a source text. The writing trait for this item defines this proficiency as the writer's ability to effectively develop real or imagined experiences or events using effective techniques, descriptive details, and clear event sequences based on a text that has been read. The holistic rubric in Part 3 describes levels of this proficiency through the situation and characters the writer establishes, the organization of the event sequence, the use of narrative techniques, the words that signal sequence and shifts in time or setting, precise and sensory language, the conclusion, the integration of source material, and errors in usage and conventions.

The score you assign is a claim about the level of this proficiency that the student demonstrated in this one response to this one item. Judge the response against the performance expected of a Grade 6 student as the rubric defines it. Students are responsible for language conventions learned in Grade 6 as well as in prior grades. The rubric footnote refers to a chart in Appendix A that is not included in this prompt, so do not guess its contents. Judge conventions by the Grade 6 expectations that the rubric descriptions and the anchor responses illustrate. Do not apply expectations from a higher or lower grade, and do not reward or penalize features that lie outside the trait.

PART 2. TASK MODEL

This part describes the task that elicited the evidence and the work product that the task produced.

The presentation material has two components. The student read the source text below, which the babysitter narrates in the first person. The student then responded to the writing prompt below under standardized assessment conditions, planning on scratch paper and writing the story on an answer document.

<source_text>

(Below copy the Source Text from Prompt A)
</source_text>

<writing_question>
(Below copy the Extended Writing Question from Prompt A)
</writing_question>

The task model variables for this item are the narrative genre, a source text that serves as the stimulus, Grade 6 as the target grade, and the demand to retell the babysitting experience from the point of view of a character other than the babysitter. The work product is the student's written story, which appears inside the <response_to_score> tags in Part 7.

Interpret the response only as an answer to this task. Features that the task requires of every student, such as narrating from another character's point of view and drawing on the events of the source text, are expected. Give credit for these features only to the extent that the rubric in Part 3 describes them, and do not treat them as merit beyond the rubric.

PART 3. EVIDENCE MODEL

This part defines what counts as evidence in the work product and how that evidence becomes a score. The evidence model has two components. The evidence rules identify observable features in the work product and use them to set the value of the observable variable. The measurement model links the observable variable to the student model variable.

Evidence rules. The holistic rubric below supplies the evidence rules for this item. Each score point names the observable features that count as evidence for that score. Begin by checking the conditions listed under Score 0. A response that is blank, copied, too limited to score, illegible, incomprehensible, written in a language other than English, off topic, off task, or offensive receives a score of 0 with the matching condition code. If no condition applies, identify in the response the observable features that the rubric names for Scores 1 through 4. Then assign the score point whose description as a whole best matches that evidence. Because the rubric is holistic, a response does not need to show every feature listed under a score point to receive that score. Base the score only on the features that the rubric names, and do not apply any rule that the rubric does not state.

Measurement model. The holistic score from 0 to 4 is the only observable variable for this item. The score is reported directly as the evidence about the student model variable, so no further combination or adjustment takes place within this task.

<rubric>
(Below copy the Four-Point Holistic Rubric from Prompt A)
</rubric>

PART 4. EVIDENTIARY ARGUMENT

This part states how the evidence supports the score. The claim is the score point, which describes the level of narrative writing proficiency that the student showed on this item. The data are the observable features you identify in the response. The warrant is the rubric in Part 3, which states which pattern of features supports each score point. The backing for the warrant is the set of anchor responses in Part 5, which show how the rubric applies to responses to this same item. Build the argument in this order. Identify the evidence in the response, map the evidence to a score point with the evidence rules, and then state the rationale that links the evidence to the chosen score point.

Before you finalize the score, check whether something other than the proficiency defined in Part 1 could explain the evidence you observed. Copying is one such explanation. A response that reproduces the source text without building a narrative of its own is copied and receives Code B. A response that places source details or source dialogue inside a narrative the student built is integrating source material, and the rubric credits that integration. Length is another possible explanation. Length is not a rubric criterion, so a longer response earns a higher score only when the added text supplies more of the evidence that the rubric describes. Errors in usage and conventions need the same care. The rubric treats errors as one feature among several in each score description, so judge them against Grade 6 expectations and by the frequency and interference with meaning that each score point describes, and weigh them together with the other features.

PART 5. ANCHOR RESPONSES

This part provides the backing for the warrant. Below are five anchor responses to this same item, one at each score point from 0 to 4. For each anchor, the score shown is the score that two independent raters agreed on. The ECD analysis shows how the evidence for that score is identified and linked to the rubric, in the format that Part 6 requires. Use the anchors to calibrate how the rubric applies to this item. Score the response in Part 7 against the rubric itself, because a response can earn a score point with a combination of features that differs from the anchor at that point. Base your analysis on the features of the response in Part 7 rather than on the wording of the anchor analyses.

### Anchor 1

**ID:** 5293

**Response:**

<student_response>
as a young enterprising entrepreneur,i quit pestering my parents for money eight moths ago since march i have been helping with assorted jobs around the naborhood i stared in the spring by helping mr.arbuckle seed his garden .soon i was mowing laws and once autumn came i was to decied to seek out new employment opportunnities.
</student_response>

**ECD analysis:**
{

"score_evidence_observed": "The response reproduces the opening of the source text almost word for word, including the lines about being an entrepreneur, helping since March, seeding Mr. Arbuckle's garden, and mowing lawns. It does not adopt another character's point of view and does not construct any new narrative in response to the prompt.",
"score_point_rationale": "The rubric assigns a 0 with a condition code to flawed responses, and Code B covers copied responses. Because the writing is copied from the source text and produces no independent narrative from another character's point of view, it matches the copied condition. Therefore the response receives a 0.",
"score": 0
}

### Anchor 2

**ID:** 5550

**Response:**

<student_response>
We were playing hiding go seek with the babysitter. I was seeking first I found her in five minutes. She was hiding behind my door. It was my turn to hide so I hide in a crevice behind the dryer. She gave up after 15 minutes. I was going to get out until I got stuck. She calmed me down and got a toolbox and a broom. She made a lever out of the toolbox and broom. She used the lever to pull the dryer up so I can slip out.
</student_response>

**ECD analysis:**
{
"score_evidence_observed": "The response retells the hide and seek events in first person using short declarative sentences that report what happened. It moves from finding the babysitter to hiding, getting stuck, being calmed, and being freed with a lever, but it includes no dialogue, little description, and no character development. The writing reads as a summary of the source events with a minimal conclusion.",
"score_point_rationale": "The rubric's Score 1 describes an attempt to write a narrative that is a summary of the story and shows little or no use of dialogue, description, and pacing. Because this response summarizes the events without narrative technique or development and provides only a minimal conclusion, it best matches Score 1. Therefore the response receives a 1.",
"score": 1
}

### Anchor 3

**ID:** 5610

**Response:**

<student_response>
Hi,my name is Conner,and I'm going to tell you about the time I got stuck in between the washer and dryer with my babysitter! It all started with a simple card game or two. We finished playing cards and decided to play hide and seek. I counted first. Don't tell him I said this,but he didn't hide very well. I mean he could have went somewhere else besides behind the door! But anyways,it was my turn,and I decided to hide where nobody would ever find me. In between the washer and dryer!He searched the whole entire house,even the garage! Finally, he gave up and when I tried to go get him,I couldn't move! I was stuck! I couldn't help myself as I started to scream and cry. Then my superhero babysitter came to the rescue! He told me to calm down,and I finally did. He was able to get me out by lifting the dryer away from me with a mop. I jumped into his arms,thanked him for saving me,and gave him a big bear hug.
</student_response>

**ECD analysis:**
{
  "score_evidence_observed": "The response is written from Connor's point of view and opens with a direct address, 'Hi, my name is Conner,' that establishes the narrator and the situation. It moves through the card game, the hide and seek, getting stuck, and the rescue, and it shows some voice such as the aside 'Don't tell him I said this, but he didn't hide very well.' The dialogue is mostly reported rather than shown, the description is thin, and the events are compressed into a brief telling with a short conclusion.",
  "score_point_rationale": "The rubric's Score 2 describes an incomplete or oversimplified narrative that introduces a situation and at least one character, organizes events in a sequence with some gaps, and attempts a narrative technique. Because this response covers the sequence from Connor's point of view but develops it in oversimplified form with limited dialogue and thin description and a weak conclusion, it best matches Score 2. Therefore the response receives a 2.",
  "score": 2
}

### Anchor 4

**ID:** 5279

**Response:**

<student_response>
"I love playing hide and seek with my babysitter " I thought to myself as I peeked into the laundry room trying to find her . "Nope not in there " I said aloud . I took of up the stairs to my room to try and find her up there " Boo!" I yelled as I flung the door open to find her there . "AH" she yelled in surprise , she was so sad I found her , I'm the best at this game . "your turn to count !" I yelled to her from the bottom of the stairs . "I'm so fast , she is a slow poke " I thought to myself as she reached the last step . I ran to go hide as she put her face up to the wall , I stumbled into the laundry room , forgot their was a step there . I squeezed myself into the middle of the washer and dryer ," this was a perfect hiding spot " I thought to myself . I waited and waited for her to find me , it took her a million years to finally give up , I knew I was to go for her to find me , I tried to push my self out of the middle of the machines when all of a sudden I relies ed that I could not do that , I pushed harder again , nothing mover other than the dryer . " help !" I screamed to the babysitter " I'm st stu stuuucckk !" I yelled out loud . The baby sitter ran to the garage and came back quickly , she wedged the broom somehow to get it to lift up so I could get out , when I got out I have her a huge hug , she basically saved my life . I love my new babysitter .............
</student_response>

**ECD analysis:**
{
  "score_evidence_observed": "The response is written from Connor's point of view and develops the experience with dialogue, internal thought, and description across a clear event sequence. It uses direct dialogue such as 'your turn to count!' and internal thought such as 'I'm the best at this game,' shows pacing through lines like 'I waited and waited,' and integrates source details including the babysitter's poor hiding, the washer and dryer, and the broom used as a lever. It provides a conclusion, and it contains noticeable errors such as 'I took of up the stairs' and 'nothing mover' that do not consistently interfere with meaning.",
  "score_point_rationale": "The rubric's Score 3 describes a complete narrative that uses narrative techniques such as dialogue and description to develop the events, integrates some source details, and provides an appropriate conclusion, while allowing a few minor errors. Because this response is a complete narrative from Connor's point of view that develops the events through dialogue, internal thought, and description, it satisfies the Score 3 criteria. Its language is not consistently precise and its errors are more than the top score allows, so it does not reach Score 4. Therefore the response receives a 3.",
  "score": 3
}

### Anchor 5

**ID:** 5194

**Response:**

<student_response>
"Connor, sweetie! Come meet the babysitter!" my mom's voice boomed up the stairs to my room. I hope this guy or girl is like those cool ones from "Adventures in babysitting" or something like that. They get chased by bad guys, lose the mom's car, and almost get arrested! This of course won't happen, but I hope some action does.  "Coming Mommy!" I reply while walking down the stairs. When I reach the last step I see it is guy who looks to be about 3 years older than me.  "Hey kid!" he says to me. I think he will be alright."Hi," I say shyly. "Okay, honey. We're just going to be gone for a few hours then we will come back." My mom said while kissing my forehead.  "Bye mommy! Bye Daddy!" I call to my parents on their way out. Now I turn to the babysitter, whose name I do not know and I probably never will.  "Are you ready to eat?" he asks. I nod in reply. After we eat pizza mom ordered for us, we watch some "Tom and Jerry".  After we finish that episode, "We should play hide 'n seek!" I announce.  "What a great idea! Should I count first?" my babysitter agrees.  "No I think you should hide first." I challenge.  "Okay," he surrenders, "but I'm very good at hiding." After about 3 seconds I find him. He needs to work on his hiding skills.  "My turn! My turn!" I say while bouncing excitedly. He counts while I hide in my favorite spot, between the washer and dryer. I thought I saw him peeking, but that option closed when he went in the completely wrong direction! It was so funny I couldn't help my giggles. After what seems like an hour he gives up. Instead of coming out laughing, I stay put and burst into a symphony of tears and screams.  He comes running and says with worry, "What's wrong? What's wrong?"  "I'm, I think I, I'm stuck." I sputter through tears. He tells me I have to calm down before he can fix it. So, slowly, but surely, I stop crying.  "I have an idea stay right here." he says. Like I'm going anywhere.  He comes back a couple minutes later with some stuff. I have no idea what is, but anything to get me out of here. He builds some kind of pulley system or maybe it's called a lever? I really don't know, but as soon as there is enough room I silently slip out and hug him. I guess I really did get that adventure.
</student_response>

**ECD analysis:**

{
  "score_evidence_observed": "The response is a well-developed narrative from Connor's point of view. It establishes the situation with the mother calling him downstairs, introduces the parents and the babysitter through dialogue and characterization, and organizes the events in a natural sequence from the introduction through dinner, the cartoon, the hide and seek, getting stuck, and the rescue. It uses sustained dialogue, internal thought, and humor to build voice, uses figurative language such as bursting 'into a symphony of tears and screams,' integrates source details including the babysitter's poor hiding and the lever used in the rescue, and closes with a conclusion that ties back to the narrator's opening wish for an adventure. Errors are very few.",
  "score_point_rationale": "The rubric's Score 4 describes a well-developed narrative that effectively establishes a situation and characters, uses narrative techniques to develop rich experiences, uses precise and sensory language consistently, integrates source material effectively, and ends with a conclusion that follows, with very few or no errors. Because this response develops the experience richly through sustained dialogue, characterization, humor, and sensory language, integrates the source details effectively, and closes with a fitting conclusion while containing very few errors, it best matches Score 4. Therefore the response receives a 4.",
  "score": 4
}

PART 6. OUTPUT SCHEMA

Return exactly one JSON object and nothing else. Do not wrap the JSON object in Markdown. The object must contain exactly three keys in the order below, and each key records one element of the evidentiary argument in Part 4.

The key "score_evidence_observed" records the data. Write a brief description of the observable features of the response that the rubric names, including the absence of a feature when that absence matters for the score. Describe only what can be observed in the response.

When you quote short phrases from the response, use single quotation marks as the anchors do, so that the JSON stays valid.

The key "score_point_rationale" records the warrant as it applies to this response. Write a brief explanation of how the observed evidence maps to the selected score point under the evidence rules. When the response lies near the boundary of an adjacent score point, state why the adjacent point fits less well. When the score is 0, name the condition code that applies.

The key "score" records the claim. Its value must be one integer from 0 to 4.

Do not invent criteria that are not included in the rubric, and do not include any text outside the JSON object.

PART 7. RESPONSE TO SCORE

<response_to_score>
[[STUDENT_RESPONSE]]
</response_to_score>

Score the response above and return only the JSON object defined in Part 6.

**A.6 Condition F: ECD with Chain of Thought Prompt**

SCORING INSTRUCTIONS

You will score one Grade 6 student response to a narrative writing item. These instructions follow Evidence-Centered Design, which is abbreviated as ECD below. Part 1 defines the student model. Part 2 defines the task model and contains the item. Part 3 defines the evidence model and contains the scoring rubric. Part 4 states the evidentiary argument that links the evidence to the score. Part 5 provides reasoning steps. Part 6 defines the output, and Part 7 contains the response you will score.

In ECD terms, your role is the evidence identification process. You examine the student's work product, identify the observable features that the evidence rules name, and set the value of one observable variable, which is the holistic score. Read every part in order before you score the response in Part 7.

PART 1. STUDENT MODEL

This part defines who produced the response and what the score claims about the student.

The writers are Grade 6 students in the United States who responded to an English Language Arts extended constructed response item. The student model has one variable, which is proficiency in narrative writing based on a source text. The writing trait for this item defines this proficiency as the writer's ability to effectively develop real or imagined experiences or events using effective techniques, descriptive details, and clear event sequences based on a text that has been read. The holistic rubric in Part 3 describes levels of this proficiency through the situation and characters the writer establishes, the organization of the event sequence, the use of narrative techniques, the words that signal sequence and shifts in time or setting, precise and sensory language, the conclusion, the integration of source material, and errors in usage and conventions.

The score you assign is a claim about the level of this proficiency that the student demonstrated in this one response to this one item. Judge the response against the performance expected of a Grade 6 student as the rubric defines it. Students are responsible for language conventions learned in Grade 6 as well as in prior grades. The rubric footnote refers to a chart in Appendix A that is not included in this prompt, so do not guess its contents. Judge conventions by the Grade 6 expectations that the rubric descriptions illustrate. Do not apply expectations from a higher or lower grade, and do not reward or penalize features that lie outside the trait.

PART 2. TASK MODEL

This part describes the task that elicited the evidence and the work product that the task produced.

The presentation material has two components. The student read the source text below, which the babysitter narrates in the first person. The student then responded to the writing prompt below under standardized assessment conditions, planning on scratch paper and writing the story on an answer document.

<source_text>
(Below copy the Source Text from Prompt A)
</source_text>

<writing_question>
(Below copy the Extended Writing Question from Prompt A)
</writing_question>

The task model variables for this item are the narrative genre, a source text that serves as the stimulus, Grade 6 as the target grade, and the demand to retell the babysitting experience from the point of view of a character other than the babysitter. The work product is the student's written story, which appears inside the <response_to_score> tags in Part 7.

Interpret the response only as an answer to this task. Features that the task requires of every student, such as narrating from another character's point of view and drawing on the events of the source text, are expected. Give credit for these features only to the extent that the rubric in Part 3 describes them, and do not treat them as merit beyond the rubric.

PART 3. EVIDENCE MODEL

This part defines what counts as evidence in the work product and how that evidence becomes a score. The evidence model has two components. The evidence rules identify observable features in the work product and use them to set the value of the observable variable. The measurement model links the observable variable to the student model variable.

Evidence rules. The holistic rubric below supplies the evidence rules for this item. Each score point names the observable features that count as evidence for that score. Begin by checking the conditions listed under Score 0. A response that is blank, copied, too limited to score, illegible, incomprehensible, written in a language other than English, off topic, off task, or offensive receives a score of 0 with the matching condition code. If no condition applies, identify in the response the observable features that the rubric names for Scores 1 through 4. Then assign the score point whose description as a whole best matches that evidence. Because the rubric is holistic, a response does not need to show every feature listed under a score point to receive that score. Base the score only on the features that the rubric names, and do not apply any rule that the rubric does not state.

Measurement model. The holistic score from 0 to 4 is the only observable variable for this item. The score is reported directly as the evidence about the student model variable, so no further combination or adjustment takes place within this task.

<rubric>
(Below copy the Four-Point Holistic Rubric from Prompt A)
</rubric>

PART 4. EVIDENTIARY ARGUMENT

This part states how the evidence supports the score. The claim is the score point, which describes the level of narrative writing proficiency that the student showed on this item. The data are the observable features you identify in the response. The warrant is the rubric in Part 3, which states which pattern of features supports each score point. Build the argument in this order. Identify the evidence in the response, map the evidence to a score point with the evidence rules, and then state the rationale that links the evidence to the chosen score point.

Before you finalize the score, check whether something other than the proficiency defined in Part 1 could explain the evidence you observed. Copying is one such explanation. A response that reproduces the source text without building a narrative of its own is copied and receives Code B. A response that places source details or source dialogue inside a narrative the student built is integrating source material, and the rubric credits that integration. Length is another possible explanation. Length is not a rubric criterion, so a longer response earns a higher score only when the added text supplies more of the evidence that the rubric describes. Errors in usage and conventions need the same care. The rubric treats errors as one feature among several in each score description, so judge them against Grade 6 expectations and by the frequency and interference with meaning that each score point describes, and weigh them together with the other features.

PART 5. REASONING STEPS

This part provides the reasoning steps for scoring the response in Part 7.

Before assigning a score, reason through the essay’s strengths and weaknesses as follows,
Step 1: Identify the narrative and evaluate its clarity.
Step 2: Evaluate how the narrative is developed.
Step 3: Evaluate organization and coherence.
Step 4: Evaluate language use and mechanics.
Step 5: Assign a holistic score (0-4) based on your analysis.

Because Part 6 allows no text outside the JSON object, write what you find in Steps 1 through 4 in "score_evidence_observed" and your reasoning for Step 5 in "score_point_rationale", so that your reasoning comes before the score.

PART 6. OUTPUT SCHEMA

Return exactly one JSON object and nothing else. Do not wrap the JSON object in Markdown. The object must contain exactly three keys in the order below, and each key records one element of the evidentiary argument in Part 4.

The key "score_evidence_observed" records the data. Write a brief description of the observable features of the response that the rubric names, including the absence of a feature when that absence matters for the score. Describe only what can be observed in the response. When you quote short phrases from the response, use single quotation marks so that the JSON stays valid.

The key "score_point_rationale" records the warrant as it applies to this response. Write a brief explanation of how the observed evidence maps to the selected score point under the evidence rules. When the response lies near the boundary of an adjacent score point, state why the adjacent point fits less well. When the score is 0, name the condition code that applies.

The key "score" records the claim. Its value must be one integer from 0 to 4.

Do not invent criteria that are not included in the rubric, and do not include any text outside the JSON object.

PART 7. RESPONSE TO SCORE

<response_to_score>
[[STUDENT_RESPONSE]]
</response_to_score>

Score the response above and return only the JSON object defined in Part 6.

**B. Confusion Matrices by Grade**

Tables B1 to B3 below report the confusion matrices for each grade that underlie Figure 4 in the main text. Figure 4 shows how the LLM classified responses within each human score level across grades. The tables in this appendix keep the original counts for each grade and add the percentage of total, which is each count divided by the 500 responses scored in that grade. The upper block presents the three prompting strategies without ECD guidance, and the lower block presents the same strategies with ECD guidance. Within each panel, rows are human scores and columns are LLM-assigned scores. As an example, the first cell of the Zero-Shot panel in Table A1 shows that 30 Grade 6 responses received a human score of 0 and an LLM-assigned score of 0, which equals 6.0% of the 500 Grade 6 responses. The cells off the diagonal show where the two scores disagree. Cells below the diagonal count responses that the LLM scored lower than the human benchmark, and cells above the diagonal count responses that the LLM scored higher.

**Table B1.** *Confusion matrices with counts and percentages of total for six prompt conditions at Grade 6*

| | LLM-Assigned Score (No ECD vs. ECD) | | | | | | | | | |
|---|---|---|---|---|---|---|---|---|---|---|
| | Zero-Shot | | | | | ECD + Zero-Shot | | | | |
| Human Score | 0 | 1 | 2 | 3 | 4 | 0 | 1 | 2 | 3 | 4 |
| 0 | **30 (6.0%)** | 51 (10.2%) | 19 (3.8%) | 0 (0.0%) | 0 (0.0%) | **34 (6.8%)** | 46 (9.2%) | 20 (4.0%) | 0 (0.0%) | 0 (0.0%) |
| 1 | 4 (0.8%) | **75 (15.0%)** | 21 (4.2%) | 0 (0.0%) | 0 (0.0%) | 7 (1.4%) | **76 (15.2%)** | 17 (3.4%) | 0 (0.0%) | 0 (0.0%) |
| 2 | 1 (0.2%) | 11 (2.2%) | **75 (15.0%)** | 13 (2.6%) | 0 (0.0%) | 0 (0.0%) | 10 (2.0%) | **82 (16.4%)** | 8 (1.6%) | 0 (0.0%) |
| 3 | 0 (0.0%) | 0 (0.0%) | 70 (14.0%) | **30 (6.0%)** | 0 (0.0%) | 0 (0.0%) | 0 (0.0%) | 56 (11.2%) | **44 (8.8%)** | 0 (0.0%) |
| 4 | 0 (0.0%) | 1 (0.2%) | 52 (10.4%) | 46 (9.2%) | **1 (0.2%)** | 0 (0.0%) | 1 (0.2%) | 46 (9.2%) | 43 (8.6%) | **10 (2.0%)** |
| | Few-Shot | | | | | ECD + Few-Shot | | | | |
| 0 | **33 (6.6%)** | 49 (9.8%) | 15 (3.0%) | 3 (0.6%) | 0 (0.0%) | **62 (12.4%)** | 23 (4.6%) | 12 (2.4%) | 3 (0.6%) | 0 (0.0%) |
| 1 | 12 (2.4%) | **65 (13.0%)** | 21 (4.2%) | 2 (0.4%) | 0 (0.0%) | 12 (2.4%) | **59 (11.8%)** | 26 (5.2%) | 3 (0.6%) | 0 (0.0%) |
| 2 | 2 (0.4%) | 11 (2.2%) | **61 (12.2%)** | 26 (5.2%) | 0 (0.0%) | 1 (0.2%) | 3 (0.6%) | **80 (16.0%)** | 16 (3.2%) | 0 (0.0%) |
| 3 | 1 (0.2%) | 0 (0.0%) | 44 (8.8%) | **55 (11.0%)** | 0 (0.0%) | 0 (0.0%) | 0 (0.0%) | 38 (7.6%) | **61 (12.2%)** | 1 (0.2%) |
| 4 | 1 (0.2%) | 0 (0.0%) | 20 (4.0%) | 77 (15.4%) | **2 (0.4%)** | 0 (0.0%) | 0 (0.0%) | 14 (2.8%) | 46 (9.2%) | **40 (8.0%)** |
| | Chain-of-Thought | | | | | ECD + Chain-of-Thought | | | | |
| 0 | **27 (5.4%)** | 57 (11.4%) | 16 (3.2%) | 0 (0.0%) | 0 (0.0%) | **27 (5.4%)** | 55 (11.0%) | 18 (3.6%) | 0 (0.0%) | 0 (0.0%) |
| 1 | 8 (1.6%) | **80 (16.0%)** | 12 (2.4%) | 0 (0.0%) | 0 (0.0%) | 7 (1.4%) | **71 (14.2%)** | 22 (4.4%) | 0 (0.0%) | 0 (0.0%) |
| 2 | 1 (0.2%) | 21 (4.2%) | **74 (14.8%)** | 4 (0.8%) | 0 (0.0%) | 0 (0.0%) | 10 (2.0%) | **86 (17.2%)** | 4 (0.8%) | 0 (0.0%) |
| 3 | 0 (0.0%) | 3 (0.6%) | 90 (18.0%) | **7 (1.4%)** | 0 (0.0%) | 0 (0.0%) | 2 (0.4%) | 69 (13.8%) | **29 (5.8%)** | 0 (0.0%) |
| 4 | 1 (0.2%) | 1 (0.2%) | 90 (18.0%) | 8 (1.6%) | **0 (0.0%)** | 0 (0.0%) | 1 (0.2%) | 51 (10.2%) | 29 (5.8%) | **19 (3.8%)** |

**Table B2.** *Confusion matrices with counts and percentages of total for six prompt conditions at Grade 7*

| | LLM-Assigned Score (No ECD vs. ECD) | | | | | | | | | |
|---|---|---|---|---|---|---|---|---|---|---|
| | Zero-Shot | | | | | ECD + Zero-Shot | | | | |
| Human Score | 0 | 1 | 2 | 3 | 4 | 0 | 1 | 2 | 3 | 4 |
| 0 | **22 (4.4%)** | 57 (11.4%) | 22 (4.4%) | 0 (0.0%) | 0 (0.0%) | **51 (10.2%)** | 31 (6.2%) | 18 (3.6%) | 1 (0.2%) | 0 (0.0%) |
| 1 | 3 (0.6%) | **52 (10.4%)** | 47 (9.4%) | 0 (0.0%) | 0 (0.0%) | 1 (0.2%) | **55 (11.0%)** | 44 (8.8%) | 2 (0.4%) | 0 (0.0%) |
| 2 | 1 (0.2%) | 5 (1.0%) | **91 (18.2%)** | 5 (1.0%) | 0 (0.0%) | 0 (0.0%) | 4 (0.8%) | **86 (17.2%)** | 12 (2.4%) | 0 (0.0%) |
| 3 | 2 (0.4%) | 0 (0.0%) | 63 (12.6%) | **36 (7.2%)** | 0 (0.0%) | 2 (0.4%) | 0 (0.0%) | 52 (10.4%) | **45 (9.0%)** | 2 (0.4%) |
| 4 | 0 (0.0%) | 0 (0.0%) | 47 (9.4%) | 22 (4.4%) | **25 (5.0%)** | 0 (0.0%) | 0 (0.0%) | 32 (6.4%) | 31 (6.2%) | **31 (6.2%)** |
| | Few-Shot | | | | | ECD + Few-Shot | | | | |
| 0 | **33 (6.6%)** | 49 (9.8%) | 18 (3.6%) | 1 (0.2%) | 0 (0.0%) | **74 (14.8%)** | 12 (2.4%) | 14 (2.8%) | 1 (0.2%) | 0 (0.0%) |
| 1 | 4 (0.8%) | **48 (9.6%)** | 47 (9.4%) | 3 (0.6%) | 0 (0.0%) | 3 (0.6%) | **65 (13.0%)** | 33 (6.6%) | 1 (0.2%) | 0 (0.0%) |
| 2 | 2 (0.4%) | 4 (0.8%) | **87 (17.4%)** | 9 (1.8%) | 0 (0.0%) | 1 (0.2%) | 0 (0.0%) | **89 (17.8%)** | 12 (2.4%) | 0 (0.0%) |
| 3 | 1 (0.2%) | 0 (0.0%) | 58 (11.6%) | **42 (8.4%)** | 0 (0.0%) | 2 (0.4%) | 0 (0.0%) | 21 (4.2%) | **77 (15.4%)** | 1 (0.2%) |
| 4 | 0 (0.0%) | 0 (0.0%) | 36 (7.2%) | 25 (5.0%) | **33 (6.6%)** | 0 (0.0%) | 0 (0.0%) | 15 (3.0%) | 30 (6.0%) | **49 (9.8%)** |
| | Chain-of-Thought | | | | | ECD + Chain-of-Thought | | | | |
| 0 | **27 (5.4%)** | 39 (7.8%) | 33 (6.6%) | 2 (0.4%) | 0 (0.0%) | **27 (5.4%)** | 34 (6.8%) | 40 (8.0%) | 0 (0.0%) | 0 (0.0%) |
| 1 | 6 (1.2%) | **42 (8.4%)** | 53 (10.6%) | 1 (0.2%) | 0 (0.0%) | 2 (0.4%) | **45 (9.0%)** | 54 (10.8%) | 1 (0.2%) | 0 (0.0%) |
| 2 | 2 (0.4%) | 10 (2.0%) | **87 (17.4%)** | 3 (0.6%) | 0 (0.0%) | 0 (0.0%) | 10 (2.0%) | **85 (17.0%)** | 7 (1.4%) | 0 (0.0%) |
| 3 | 0 (0.0%) | 3 (0.6%) | 64 (12.8%) | **34 (6.8%)** | 0 (0.0%) | 1 (0.2%) | 0 (0.0%) | 52 (10.4%) | **48 (9.6%)** | 0 (0.0%) |
| 4 | 0 (0.0%) | 0 (0.0%) | 71 (14.2%) | 3 (0.6%) | **20 (4.0%)** | 0 (0.0%) | 0 (0.0%) | 46 (9.2%) | 13 (2.6%) | **35 (7.0%)** |

**Table B3.** *Confusion matrices with counts and percentages of total for six prompt conditions at Grade 8*

| Human Score | LLM-Assigned Score (No ECD vs. ECD) | | | | | | | | | |
|---|---|---|---|---|---|---|---|---|---|---|
| | Zero-Shot | | | | | ECD + Zero-Shot | | | | |
| | 0 | 1 | 2 | 3 | 4 | 0 | 1 | 2 | 3 | 4 |
| 0 | **41 (8.2%)** | 33 (6.6%) | 24 (4.8%) | 2 (0.4%) | 0 (0.0%) | **53 (10.6%)** | 25 (5.0%) | 19 (3.8%) | 3 (0.6%) | 0 (0.0%) |
| 1 | 5 (1.0%) | **63 (12.6%)** | 31 (6.2%) | 1 (0.2%) | 0 (0.0%) | 1 (0.2%) | **73 (14.6%)** | 22 (4.4%) | 4 (0.8%) | 0 (0.0%) |
| 2 | 0 (0.0%) | 3 (0.6%) | **91 (18.2%)** | 6 (1.2%) | 0 (0.0%) | 0 (0.0%) | 0 (0.0%) | **86 (17.2%)** | 14 (2.8%) | 0 (0.0%) |
| 3 | 1 (0.2%) | 0 (0.0%) | 54 (10.8%) | **45 (9.0%)** | 0 (0.0%) | 2 (0.4%) | 0 (0.0%) | 39 (7.8%) | **59 (11.8%)** | 0 (0.0%) |
| 4 | 2 (0.4%) | 0 (0.0%) | 47 (9.4%) | 39 (7.8%) | **12 (2.4%)** | 1 (0.2%) | 0 (0.0%) | 37 (7.4%) | 45 (9.0%) | **17 (3.4%)** |
| | Few-Shot | | | | | ECD + Few-Shot | | | | |
| 0 | **51 (10.2%)** | 25 (5.0%) | 24 (4.8%) | 0 (0.0%) | 0 (0.0%) | **81 (16.2%)** | 6 (1.2%) | 10 (2.0%) | 3 (0.6%) | 0 (0.0%) |
| 1 | 3 (0.6%) | **60 (12.0%)** | 35 (7.0%) | 2 (0.4%) | 0 (0.0%) | 2 (0.4%) | **81 (16.2%)** | 16 (3.2%) | 1 (0.2%) | 0 (0.0%) |
| 2 | 0 (0.0%) | 10 (2.0%) | **82 (16.4%)** | 8 (1.6%) | 0 (0.0%) | 0 (0.0%) | 1 (0.2%) | **86 (17.2%)** | 12 (2.4%) | 1 (0.2%) |
| 3 | 2 (0.4%) | 0 (0.0%) | 43 (8.6%) | **54 (10.8%)** | 1 (0.2%) | 0 (0.0%) | 0 (0.0%) | 16 (3.2%) | **83 (16.6%)** | 1 (0.2%) |
| 4 | 0 (0.0%) | 0 (0.0%) | 36 (7.2%) | 46 (9.2%) | **18 (3.6%)** | 0 (0.0%) | 0 (0.0%) | 13 (2.6%) | 25 (5.0%) | **62 (12.4%)** |
| | Chain-of-Thought | | | | | ECD + Chain-of-Thought | | | | |
| 0 | **29 (5.8%)** | 45 (9.0%) | 26 (5.2%) | 0 (0.0%) | 0 (0.0%) | **54 (10.8%)** | 20 (4.0%) | 23 (4.6%) | 3 (0.6%) | 0 (0.0%) |
| 1 | 5 (1.0%) | **65 (13.0%)** | 30 (6.0%) | 0 (0.0%) | 0 (0.0%) | 2 (0.4%) | **70 (14.0%)** | 27 (5.4%) | 1 (0.2%) | 0 (0.0%) |
| 2 | 0 (0.0%) | 9 (1.8%) | **88 (17.6%)** | 3 (0.6%) | 0 (0.0%) | 0 (0.0%) | 4 (0.8%) | **91 (18.2%)** | 5 (1.0%) | 0 (0.0%) |
| 3 | 1 (0.2%) | 3 (0.6%) | 69 (13.8%) | **27 (5.4%)** | 0 (0.0%) | 1 (0.2%) | 0 (0.0%) | 49 (9.8%) | **50 (10.0%)** | 0 (0.0%) |
| 4 | 0 (0.0%) | 1 (0.2%) | 78 (15.6%) | 7 (1.4%) | **14 (2.8%)** | 1 (0.2%) | 1 (0.2%) | 54 (10.8%) | 17 (3.4%) | **27 (5.4%)** |